%% file: main.tex
\documentclass[10pt,journal,compsoc]{IEEEtran}

\usepackage{graphicx}
\usepackage{amsmath}
\usepackage{booktabs}
\usepackage{multirow}
\usepackage{xcolor}

\usepackage{caption}
\usepackage{marvosym}

\ifCLASSOPTIONcompsoc
  \usepackage[nocompress]{cite}
\else
  \usepackage{cite}
\fi
\ifCLASSINFOpdf
\else
\fi
\begin{document}
%
\title{Task-Aware Sampling Layer for Point-Wise Analysis}
%
%
%
%

\author{Yiqun Lin, Lichang Chen, Haibin Huang, Chongyang Ma,\\ Xiaoguang Han\textsuperscript{\Letter},~\IEEEmembership{Member,~IEEE,} Shuguang Cui,~\IEEEmembership{Fellow,~IEEE}
\IEEEcompsocitemizethanks{\IEEEcompsocthanksitem 
Y. Lin, X. Han and S. Cui are with the School of Science and Engineering, The Chinese University of Hong Kong, Shenzhen, China.
\protect\\
E-mail: \{yiqunlin@link., hanxiaoguang@, shuguangcui@\}cuhk.edu.cn
\IEEEcompsocthanksitem L. Chen is with the Department of Electrical and Computer Engineering, University of Pittsburgh, PA, USA.
\protect\\
E-mail: bobchen23@pitt.edu
\IEEEcompsocthanksitem H. Huang and C. Ma are with Kuaishou Technology.
\protect\\
E-mail: \{jackiehuanghaibin, chongyangm\}@gmail.com
\IEEEcompsocthanksitem Xiaoguang Han is the corresponding author.
}
\thanks{Manuscript received January 30, 2022; revised March 23, 2022.}
}

%
%

\markboth{IEEE TRANSACTIONS ON VISUALIZATION AND COMPUTER GRAPHICS,~Vol.~XX, No.~X, XXXX~2022}{Yiqun \MakeLowercase{\textit{et al.}}: Task-Aware Sampling Layer for Point-Wise Analysis}

\newcommand{\yq}[1]{{\color[rgb]{0.0, 0.0, 0.0}{#1}}}
\newcommand{\yqq}[1]{{\color[rgb]{0.0, 0.0, 0.0}{#1}}}
\newcommand{\red}[1]{{\color[rgb]{0.0, 0.0, 0.0}{#1}}}
\newcommand{\green}[1]{{\color[rgb]{0.0, 0.0, 0.0}{#1}}}
\newcommand{\cmt}[1]{{\color[rgb]{0.0, 0.0, 0.0}{#1}}}

\IEEEtitleabstractindextext{%
\begin{abstract}
Sampling, grouping, and aggregation are three important components in the multi-scale analysis of point clouds. In this paper, we present a novel data-driven sampler learning strategy for point-wise analysis tasks. Unlike the widely used sampling technique, Farthest Point Sampling (FPS), we propose to learn sampling and downstream applications jointly. Our key insight is that uniform sampling methods like FPS are not always optimal for different tasks: sampling more points around boundary areas can make the point-wise classification easier for segmentation. Towards this end, we propose a novel sampler learning strategy that learns sampling point displacement supervised by task-related ground truth information and can be trained jointly with the underlying tasks. We further demonstrate our methods in various point-wise analysis tasks, including semantic part segmentation, point cloud completion, and keypoint detection. Our experiments show that jointly learning of the sampler and task brings better performance than using FPS in various point-based networks.
\end{abstract}

}

\maketitle

\IEEEdisplaynontitleabstractindextext

%
\IEEEpeerreviewmaketitle

\newcommand{\loss}{\mathcal{L}}

\input{subfiles/sec1_introduction.tex}

\input{subfiles/sec2_related_works.tex}
\input{subfiles/sec3_method.tex}
\input{subfiles/sec4_experiments.tex}

\input{subfiles/sec5_ablation.tex}

\input{subfiles/sec6_discussion.tex}
\input{subfiles/sec7_conclusion.tex}

\ifCLASSOPTIONcaptionsoff
  \newpage
\fi

\bibliographystyle{IEEEtran}
\bibliography{egbib}

\end{document}

%% file: subfiles/sec1_introduction.tex
\IEEEraisesectionheading{
\section{Introduction}\label{sec:introduction}}

\IEEEPARstart{W}ITH the rapid development of 3D sensing devices and the growing number of 3D shape repositories available online, it has become easier to access and process 3D data. One of the remaining challenges is 3D point cloud analysis, a popular 3D representation wildly used in the areas of robotics, autonomous driving, and virtual reality applications \cite{pomares2018ground, yue2018lidar, behl2017bounding, rambach2017poster}. 
\yq{In deep networks~\cite{he2016deep}, pooling layers (e.g., max-pooling and avg-pooling) play an important role in expanding receptive fields and reducing the computational cost. Similarly, in point-based networks, point sampling is used to downscale point clouds by selecting a subset of points. Sampled points are regarded as centroids to gather features of neighbor points from the original point cloud~\cite{qi2017pointnet++}. The most commonly used sampling strategy in point-based networks is farthest point sampling (FPS)~\cite{eldar_fps} that samples points uniformly. However, it does not consider prior semantic knowledge.}

\begin{figure}[t]
\centering
\includegraphics[width=\linewidth]{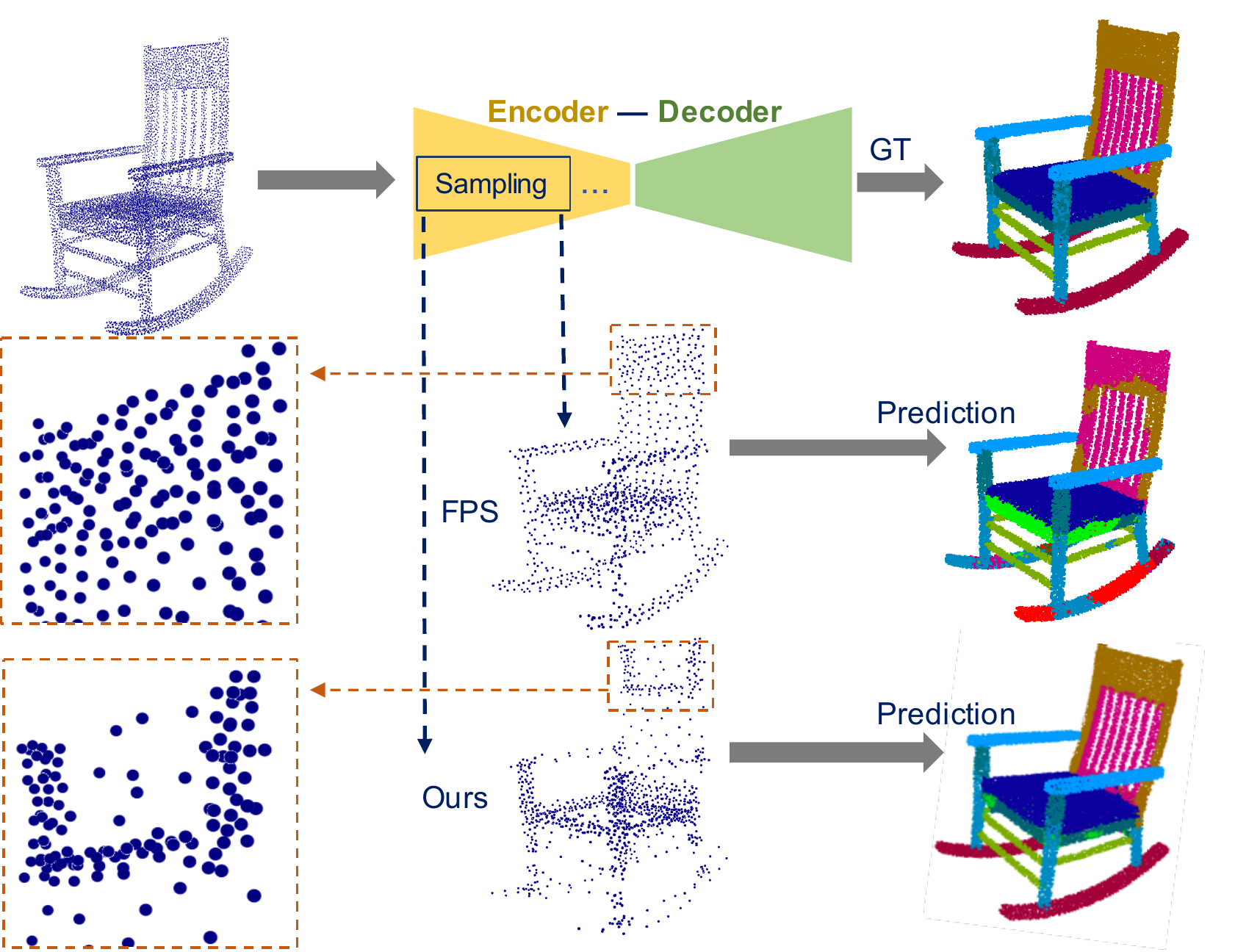}
\caption{We explore task-related samplings that can be learned by a novel training strategy in point-wise analysis tasks. For example, sampling points at a higher rate in boundary areas of a point cloud can help to aggregate more detailed features in boundary areas and significantly improve semantic segmentation results.}
\label{fig:teaser}
\end{figure}

\yq{In point-based tasks, local details carry significant information (e.g., semantics and textures), and the sampling can affect local feature aggregation via changing neighbor points.}
Our observation is that sampled points that have different semantics based on downstream tasks can \yq{improve local feature learning and further boost the performance.}
Taking part segmentation as an example, points around the segmentation boundary are more sensitive than those that are far away. Hence the prior requires a higher sampling rate to group more boundary points for better results. 
\yq{However, for some tasks that mainly rely on global features, such as classification, there is little difference between different sampling strategies since global features are not sensitive to local differences, as discussed in Section~\ref{sec:dis_cls_seg}.}

\yq{Hence, in this paper, we explore the sampling in point-wise analysis tasks.}
In general, it is always possible to design a handcrafted sampler with ground-truth (GT) information that is related to the task to provide better performance than FPS. Here we call the handcrafted sampler involving task-related GT information as \yq{task-aware sampler}. Since the GT information cannot be observed during inference, our goal is to learn that task-aware sampler for each specific task automatically.

To this end, we developed a supervised sampler learning strategy. The core idea is to jointly learn the sampling and optimize the given task. Specifically, the network learns to minimize the displacement of sampled points w.r.t task-aware sampler and the downstream task-related loss. After training, the learned sampler can provide task-related information for better inference by approximating the task-aware sampling points. To verify the proposed strategy,  we integrate it into different point-wise learing architectures such as PointNet++ \cite{qi2017pointnet++}, PointConv \cite{wu2019pointconv}, FPConv \cite{lin2020fpconv} and evaluate them on various tasks, including segmentation, completion, and keypoint detection. Our experiments show that jointly training the sampler and tasks brings a significant improvement over previous methods with FPS.

To summarize, the main contributions of this work include:
\begin{itemize} 
    \item We explore sampling in point-wise analysis tasks and propose a novel task-related sampler learning strategy.
    \item The proposed sampler can be integrated into different point-wise architectures such as PointNet++ \cite{qi2017pointnet++}, PointConv \cite{wu2019pointconv}, FPConv \cite{lin2020fpconv}, \yq{DGCNN~\cite{dgcnn}, and PointCNN~\cite{li2018pointcnn}.}
    \item Extensive experiments are conducted on various tasks, including segmentation, completion, and keypoint detection, to show \yqq{the improvement} of our learnable sampler compared to FPS.
\end{itemize}

%% file: subfiles/sec2_related_works.tex
\section{Related Works}\label{sec:related_works}

\noindent
\textbf{Deep learning on point clouds.}
Deep learning has developed rapidly in recent years. In 2D images, convolutional neural networks (CNNs) play a significant role and greatly improve performance on almost every vision task. Nevertheless, due to the irregularity and sparsity of point clouds, applying pooling and convolution layers on point clouds is much more difficult than applying them on 2D images. Comprising approaches are proposed like applying convolution on 3D voxel space. Specifically, earlier works such as \cite{wu20153d, maturana2015voxnet, qi2016volumetric} voxelized the point cloud into volumetric grids, and apply 3D convolution and pooling layers for feature extraction. However, they suffer from low resolution and high computational cost problems, since the point cloud is extremely sparse and a large number of grids are empty. More recently, \cite{spconv2019, SubmanifoldSparseConvNet, choy20194d} proposed to apply 3D convolution in a sparse way to reduce the computational cost, but their method will still lose details during voxelization.

To process point clouds efficiently and reduce detail loss, recent works explore new designs of local aggregation operators on point clouds. PointNet \cite{qi2017pointnet} proposed to use a shared MLP followed by a max-pooling layer to approximate a continuous set function for unordered data processing. \cite{hua2018pointwise, xu2018spidercnn, groh2018flex, atzmon2018point, thomas2019kpconv, wu2019pointconv} defined convolution kernels for points with learnable weights, which are similar to image convolution. When the relationships among points have been established, graph convolution \cite{dgcnn, xu2020gridgcn, wang2018local, te2018rgcnn, li2019deepgcns} can then be applied for local feature learning with high efficiency. Since 3D sensing devices capture points on object surfaces, \cite{lin2020fpconv, tatarchenko2018tangent, huang2019texturenet, yang2020pfcnn} attempted to operate feature aggregation on surfaces directly and apply 2D CNN or surface CNN instead of in a volumetric way.
Previous aggregation methods are widely used in various high-level point cloud analysis architectures, including segmentation \cite{yu2019partnet, thomas2019kpconv, lin2020fpconv, dai2017scannet}, point cloud completion \cite{gong2021me, nie2020skeleton, occnet2019, yin2018p2pnet}, object detection \cite{qi2017frustum, qi2019deep, Shi_2019_CVPR, shi2020pv} and keypoint detection \cite{you2020keypointnet}. However, those works mainly focus on local feature learning and architecture design. Few researchers study the impact of sampling in point-wise analysis, although it is important and indispensable.

\vspace{0.8em} \noindent
\yq{\textbf{Point sampling for sparse representation.} Sampling as a task of selecting a subset of points to represent the original one at a sparse scale, can improve the efficiency of 3D data processing and has broad applications. 
So far, there are various handcrafted sampling strategies, such as farthest point sampling (FPS)~\cite{eldar_fps} to select points that are farthest away from each other, and grid (voxel) sampling~\cite{thomas2019kpconv, spconv2019} to downsample over 3D voxels. 
Recent works S-NET~\cite{dovrat2018learning_to_sample} and SampleNet~\cite{lang2020samplenet} developed parametric and task-oriented sampling strategies for a better sparse representation than FPS, demonstrating that FPS is not the best way for sparse encoding in some specific tasks, including classification, registration, and reconstruction. They guided the sampler to learn a similar point distribution under the supervision of the original point cloud and optimize it for downstream tasks.}



\vspace{0.8em} \noindent
\textbf{Point sampling for point cloud analysis.} In point cloud analysis, sampling also plays a key role in expanding receptive fields and reducing the computational cost. In multiscale analysis architectures like PointNet++~\cite{qi2017pointnet++} and DGCNN~\cite{dgcnn}, rather than as an input like \cite{dovrat2018learning_to_sample, lang2020samplenet}, sampling is similar to pooling layers, and the sampled points are used as centroids to group neighbor points and aggregate features.
Widely used sampling strategies incorporate FPS~\cite{qi2017pointnet++} and grid (voxel) sampling~\cite{thomas2019kpconv, spconv2019}. Recently, \cite{xu2020gridgcn} proposed a coverage-aware grid query (CAGQ) module for seeking optimal coverage over the original point cloud. 

Those samplers mentioned above are non-parametric and intuitive. \yq{Some recent works proposed to adopt adaptively learnable samplers for better robustness and performance.}
%
\yq{PointFilter~\cite{zhang2020pointfilter} proposed to learn point-wise displacement for removing point noise;} PointASNL~\cite{yan2020pointasnl} proposed a local adaptive shifting targeting on sampled points to improve the robustness and reduce the sensitivity to noisy points. 
\yq{Although S-NET \cite{dovrat2018learning_to_sample} and SampleNet \cite{lang2020samplenet} are used for point sampling, they bring negligible improvement over FPS in point-wise analysis tasks.
The possible reason is that the sampled points are still uniformly distributed as they are under the supervision of the original point cloud and have no local differences with FPS. We will discuss this problem in-depth in Section~\ref{sec:dis_samplenet}.} Hence, we propose a novel supervised sampler learning strategy to learn point sampling of different point distributions in point-wise architectures.



%% file: subfiles/sec3_method.tex
\section{Method}

We describe our sampler learning strategy in this section. We first revisit the impact of sampling on point-based networks and demonstrate that different tasks require different sampling strategies. We then present our learning approach for task-oriented sampling, from supervision generation to jointly training.

\begin{figure}[t]
\centering
\includegraphics[width=1\linewidth]{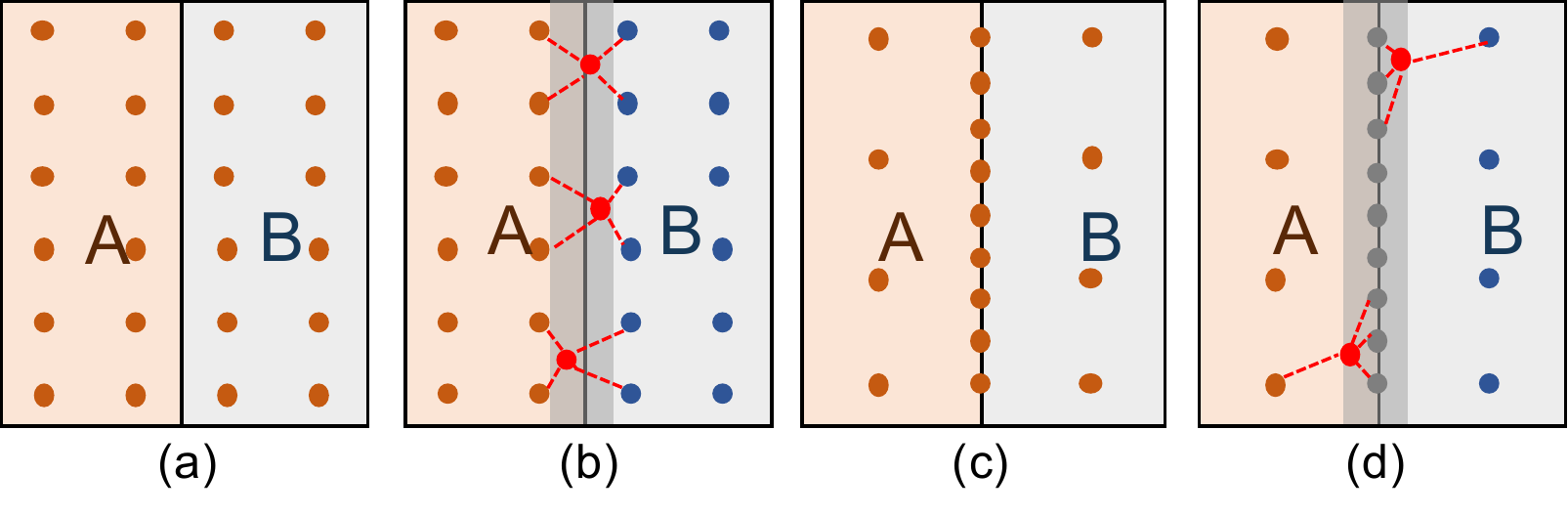}
\caption{Comparison of uniform sampling and task-aware sampling. A plane is split to two parts, A and B. Points sampled uniformly in (a). As shown in (b), in upsampling steps, for points close to boundary area, their neighbors are half in part A and half in part B, which will increase the difficulty of classification. While in (c) and (d), points are sampled densely in boundary area, then the neighbors are only from the boundary area and part A (or B).}
\label{fig:non-uniform}
\end{figure}

\subsection{Sampling in \yq{Point-based Networks}} \label{sec:task_sampling}

\yq{Sampling is widely used in multi-scale point-based networks. We take PointNet++~\cite{qi2017pointnet++} for demonstrating the importance of point sampling. The basic feature extraction module is set abstraction in PointNet++, which consists of sampling, grouping, and PointNet~\cite{qi2017pointnet} layer. Sampling layer aims to select a subset of points from input points, and grouping layer is applied to each sampled point for querying neighbor points from input points in the Euclidean space. For each sampled point, a PointNet layer is further adopted to gather features from its neighbor points. In feature propagation (upsampling) layers of PointNet++, to recover the size of the point cloud, feature values of input points are interpolated from sampled points by inverse distance weighted average based on $k$ nearest neighbors.}


Usually, the sampling algorithm is predefined (more like a hyperparameter) for point-based networks and it is task-independent. Among existing sampling methods, FPS is a uniform sampling strategy and can cover all points in space as much as possible and give a stable performance on different tasks \cite{qi2017pointnet++, wu2019pointconv, lin2020fpconv}. However, FPS is agnostic to downstream applications, which means the sampled points are selected only based on low-level information without considering object semantic and task-related information. The underlying assumption is that each point has the same importance to task learning.



Following this direction, we study the sampling problem in point-wise analysis. Our observation is that point-wise analysis tasks like point labeling would have significant point-wise semantics, and FPS is not optimal to select representative samples to learn. Taking point labeling as an example, as shown in Fig.~\ref{fig:non-uniform}, points near the segmentation boundary can provide more information than points far away. A plane is split evenly into two parts, A and B. During feature learning, methods like FPS will generate uniformly distributed samples like those in (a). The next step is to propagate and interpolate the learned features, as shown in (b). For points close to the boundary, their neighbor points can be two from part A and two from part B, increasing the difficulty of classifying which part they belong to since they may be confused when gathering two-part features. Hence, if we use a non-uniform sampling method like (c), adding more sampling points around the boundary can significantly improve the learning results (d) for the reason that the neighbors are only from one part and boundary area.

\begin{figure}[t]
\centering
\includegraphics[width=1\linewidth]{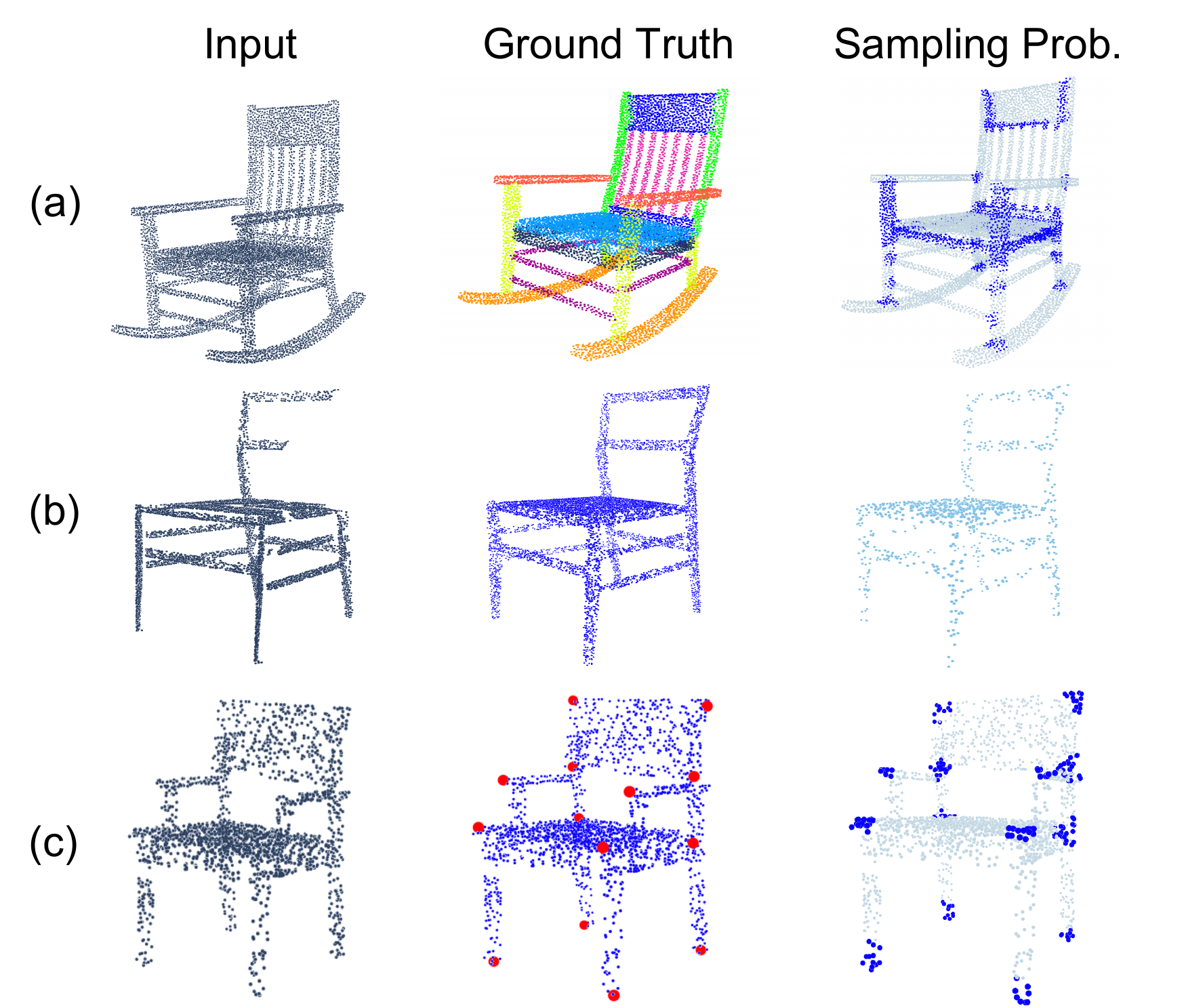}
\caption{Task-aware sampling points in different tasks. (a) In semantic segmentation task, points are sampled at a higher sampling rate in boundary areas while at a lower sampling rate in other areas; (b) in point cloud completion task, we directly perform uniform sampling on the complete point cloud (ground truth); (c) in keypoint detection task, points near keypoints \yq{(points in red)} have a higher sampling rate than others.}
\label{fig:pre_designed}
\end{figure}

\subsection{Learning the Sampling} \label{sec:learn_sampling}
As demonstrated in Section \ref{sec:task_sampling}, introducing task-related information to the sampling stage can boost performance. However, the challenge is that there is no such information during the inference time to rely on. We propose a novel sampling learning strategy in a data-driven fashion: we first design a sampling method involving task information for each task, then use them as an additional constraint during the task learning. Specifically, we refer to the task-related sampling as task-aware sampling, and jointly train the sampler and the underlying task so that the generated sample points can approximate the task-aware sampling points and optimize the task loss.

\begin{figure*}[t]
\centering
\includegraphics[width=1\linewidth]{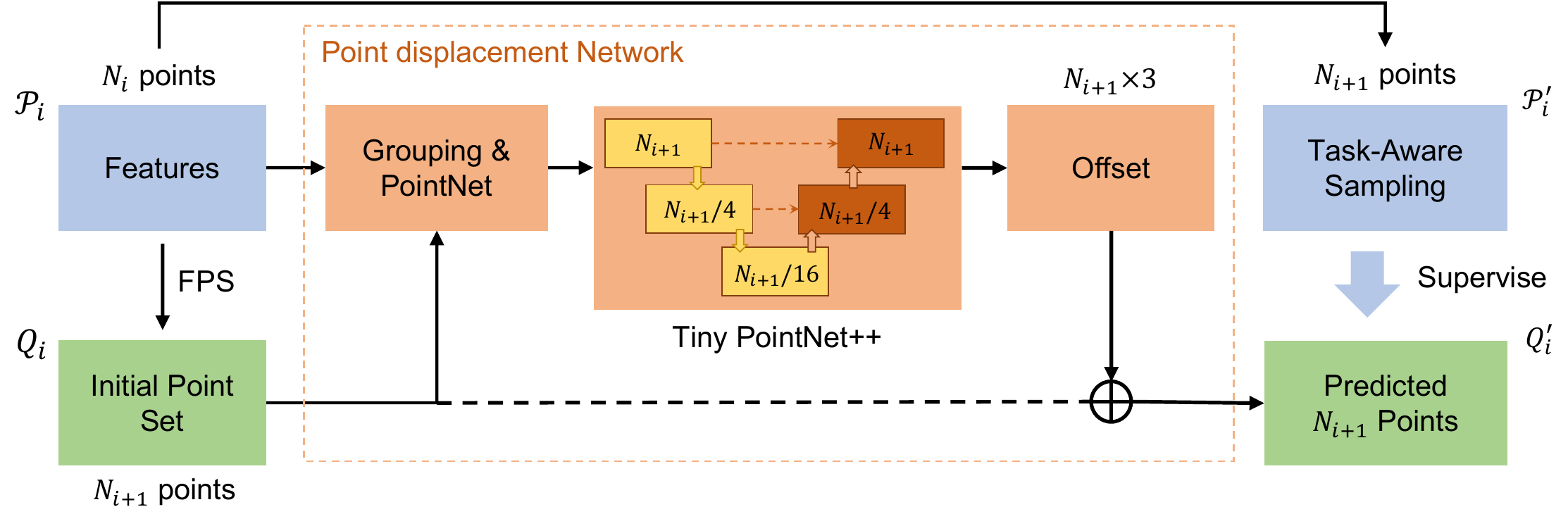}
\caption{Learn point displacement. Given a point set $\mathcal{P}_i$ with features and an initial subset $Q_i$ \yq{sampled by FPS from $\mathcal{P}_i$, a tiny version of PointNet++ with two downsampling layers} is used for point-wise feature learning. We predict point-wise offsets to $Q_i$ or coordinates directly as the final sampled points. \yq{Task-aware sampling is designed based on specific downstream task to generate supervision points from input point set $\mathcal{P}_i'$.}}
\label{fig:sampler_pipeline}
\end{figure*}

\vspace{0.8em}
\noindent \textbf{Supervising sampling.}
We first introduce how to incorporate task information into sampling. It is a process of adaptive sampling where we can change the underlying point density. Specifically, we change the sampling rate of different areas w.r.t the given task. \yq{To formulate, let the input point set be $\mathcal{P}_i$ with $N_i$ points, we firstly design a task-aware sampling function $g(\cdot)$ based on the downstream task and apply $g(\cdot)$ on $\mathcal{P}_i$ to obtain task-aware sampling point set $\mathcal{P}_i'$ with $N_{i+1}$ points:
\begin{equation}
    \mathcal{P}_i' = g(\mathcal{P}_i; N_{i+1}).
\end{equation}}
In this paper, we demonstrate three different tasks, including segmentation, completion, and keypoint detection. As shown in Fig.~\ref{fig:pre_designed}, in the segmentation task, since the points in boundary areas are more sensitive, a higher sampling rate in boundary areas should be helpful for predictions on these points. Similarly, in point cloud completion, instead of sampling points from the original partial point cloud, we uniformly sample from the complete point cloud (ground truth) for a stronger hint for better surface completion. Sampling more points around keypoints will also be useful for keypoint detection. For more implementation details of the task-aware sampler design, please refer to Section \ref{sec:exp}. \yq{Task-aware sampling points are generated before training for saving time since the sampling points are only relevant to the data.}

\vspace{1.5em}
\noindent \textbf{Learning points displacement.} 
Although we have adaptive sampling methods to generate samples related to task information, it is still hard to directly apply to inference time without ground truth information. 
A naive approach is to train a network to learn such sampling in an end-to-end manner by generating sufficient paired data of point clouds and sample sets. However, \yq{our experiments in Section~\ref{sec:abla} (``training strategy'') show that} such training can not recover the underlying task-related semantic information and may lead to overfitting of geometry-related features. 
Instead, we propose to train the sampler and task jointly by adding positional constraints of samples during training. To be specific, we additionally learn point displacement to approximate the distribution of the task-aware sampling. 
%
As shown in Fig.~\ref{fig:sampler_pipeline}, we firstly select $N_{i+1}$ points \yq{by FPS from input point set $\mathcal{P}_i$} as initial point set $Q$, and group point features of $Q$ from $\mathcal{P}_i$. Then a successive \yq{tiny} PointNet++ layer is used for point-wise offsets learning or directly regressing coordinates \yq{to generate sampling points $Q_i'$. Then the task-aware sampling points $\mathcal{P}_i'$ is used to supervise $Q_i'$ and guide the displacement network to learn similar sampling distribution.}
For the selection of learning ways: offset learning or coordinate regression, we conduct ablative experiments in Section \ref{sec:disp_learn}.

\begin{figure*}[t]
\centering
\includegraphics[width=1\linewidth]{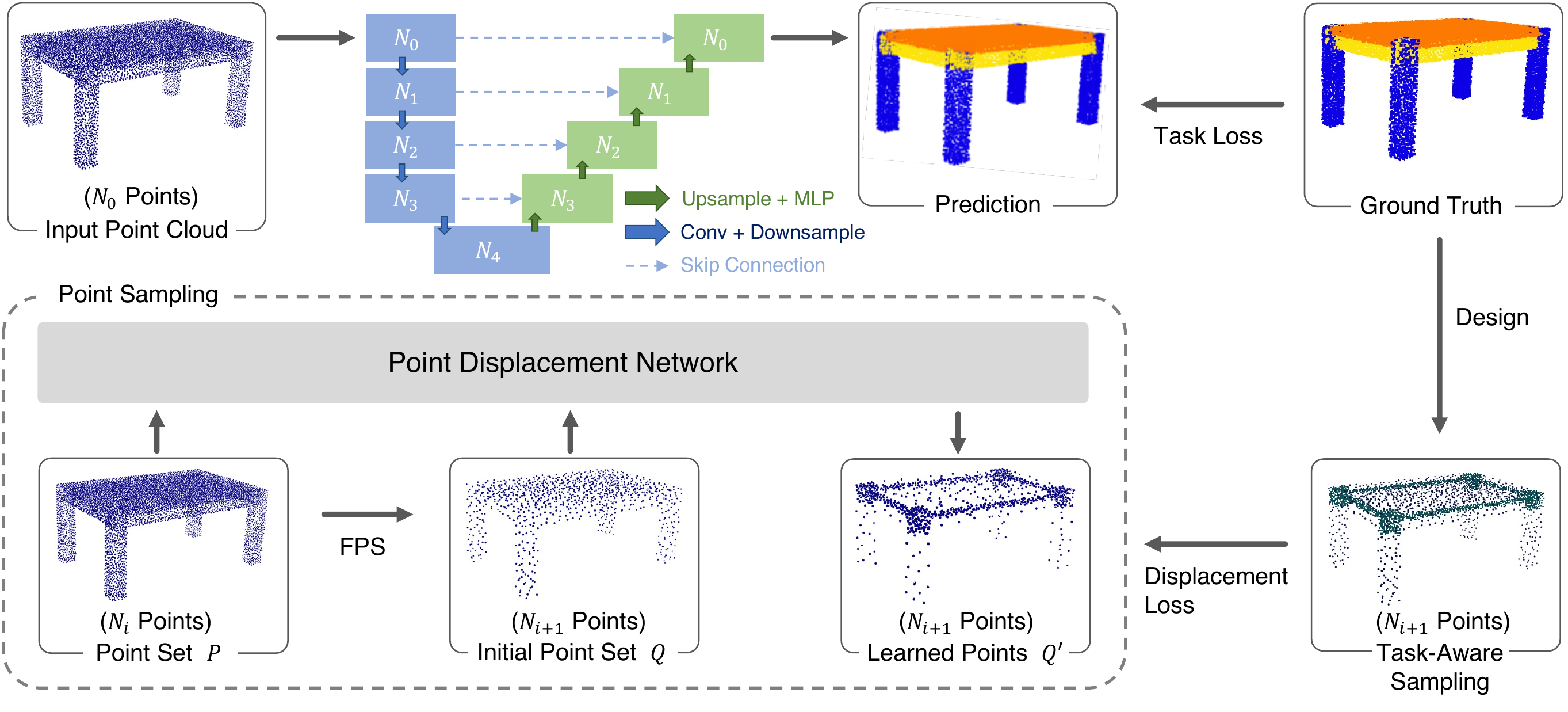}
\caption{Network architecture for point-wise analysis. We use a U-Net like architecture that consists of four downsampling and four upsampling layers. The final output is supervised by ground truth with task loss. In the point sampling step, we first select a subset of points as the initial point set \yq{by FPS}, and then predict the point-wise offset (or coordinates) to obtain the learned sampling points\yq{; see Fig.~\ref{fig:sampler_pipeline}.} The downsampling module uses the learned sampling points to replace FPS points (baseline). The displacement network is supervised by task-aware sampling with CD (or EMD) loss, and also jointly supervised with task loss since it is one part of the whole task network.}
\label{fig:net_achitecture}
\end{figure*}

\vspace{0.8em}
\noindent \textbf{Network architecture.}
For point-wise analysis, \yq{the U-Net~\cite{ronneberger2015u} like} network architecture is shown in Fig.~\ref{fig:net_achitecture}, which includes four downsampling layers, four upsampling layers, and a fully connected layer at the end for point-wise prediction. 
\yq{Similar to the set abstraction (SA) module in PointNet++~\cite{qi2017pointnet++}, the downsampling layer consists of point sampling, neighbor point grouping, and local feature aggregation. FPS is used for sampling in the baseline model, and we replace it with the proposed sampler for comparison. We conduct experiments with different feature aggregation operators including PointNet~\cite{qi2017pointnet}, PointConv~\cite{wu2019pointconv}, FPConv~\cite{lin2020fpconv}, PointCNN~\cite{li2018pointcnn}, and DGCNN~\cite{dgcnn} to demonstrate the effectiveness of our proposed sampler in the following Section. Feature propagation (FP) module~\cite{qi2017pointnet++} is used in upsampling layers.}


\vspace{0.8em}
\noindent \textbf{Loss functions.}
To put the predicted points closer to the sampled points of task-aware sampler, we propose to optimize the displacement loss $\loss_{\text{disp}}$ using Chamfer distance (CD) \cite{huang2020pfnet} or the earth mover's distance (EMD) \cite{yuan2018pcn}, which are given by
\begin{equation}
\begin{split}
    \loss_{\text{CD}} (Q_i', \mathcal{P}_i') =& \frac{1}{2} (\sum_{p \in Q_i'} \min_{q \in \mathcal{P}_i'} ||p - q||_2 \\
                                 & + \sum_{q \in \mathcal{P}_i'} \min_{p \in Q_i'} ||q - p||_2),
\end{split}
\label{eq:cd}
\end{equation}
\begin{equation}
\begin{split}
    \loss_{\text{EMD}} (Q_i', \mathcal{P}_i') = \min_{\phi: Q_i' \rightarrow \mathcal{P}_i'} \sum_{p \in Q_i'} ||p - \phi(p)||_2,
\end{split}
\label{eq:emd}
\end{equation}
where $Q_i'$ is the predicted point set, $\mathcal{P}_i'$ is the point set sampled by task-aware sampler and $\phi$ indicates the bijection between $Q_i'$ and $\mathcal{P}_i'$. 
\yq{For EMD supervision with predicted points, the bijection matching changes during training, which can make the training process unstable. Therfore, }
we recommend to compute bijection matching using the initial point set instead of predicted points for a stable shape supervision:
%
\begin{equation}
\begin{split}
    \loss_{\text{EMD}^\star} (Q_i', &Q_i, \mathcal{P}_i') = \\ &\min_{\phi: Q_i \rightarrow \mathcal{P}_i'} \sum_{p \in Q_i', h \in Q_i} ||p - \phi(h)||_2,
\end{split}
\label{eq:emd_hint}
\end{equation}
\yq{where $Q_i$ is the initial point and $p \in Q_i'$ is the prediction of $h \in Q_i$.} We compare the impact of different displacement loss functions in Section \ref{sec:disp_loss}.
%
To optimize the predicted point set $Q_i'$ to the task, we jointly train the sampler with task loss $\loss_\text{task}$. The overall loss function $\loss_\text{total}$ is given by:
\begin{equation}
    \loss_{\text{total}} = \loss_{\text{task}} + \alpha \cdot \loss_{\text{disp}},
\label{eq:joint}
\end{equation}
where $\alpha$ indicates the scale ratio and varies among different tasks, \yq{and $\loss_\text{disp}$ is the displacement loss shown in Equation~\ref{eq:cd}-\ref{eq:emd_hint}.}

%% file: subfiles/sec4_experiments.tex
\section{Experiments} \label{sec:exp}
To demonstrate the effectiveness of our proposed sampler learning strategy, we conduct experiments with three different local aggregation operators including PointNet++ \cite{qi2017pointnet++}, PointConv \cite{wu2019pointconv}, FPConv \cite{lin2020fpconv}, \yq{DGCNN~\cite{dgcnn}, and PointCNN~\cite{li2018pointcnn}} under the same architecture shown in Fig.~\ref{fig:net_achitecture}. Comparison experiments are conducted in three point-wise analysis tasks, including semantic part segmentation on PartNet \cite{yu2019partnet}, point cloud completion on ShapeNet \cite{shapenet} and keypoint detection on KeyPointNet \cite{you2020keypointnet}. In the following experiments, we compare different sampling strategies only in the second \yq{downsampling} layer. Ablation study on layer selection is shown in Section \ref{sec:layer_selection}.

\input{subfiles/sec4_experiments/sec4_1_segmentation.tex}

\input{subfiles/sec4_experiments/sec4_2_completion.tex}
\input{subfiles/sec4_experiments/sec4_3_keypoint.tex}

%% file: subfiles/sec4_experiments/sec4_1_segmentation.tex
\subsection{Semantic Segmentation} \label{sec:exp_seg}
\noindent\textbf{Dataset.}
We conduct experiments on PartNet~\cite{yu2019partnet} dataset, which is annotated with fine-grained, instance-level, and hierarchical 3D part information. It consists of 573,585 part instances over 26,671 3D models and covers 24 object categories. \yq{Most categories are labeled in three levels of segmentation: coarse-, middle- and fine-grained.}
For some categories with a small amount of data (less than 2,000), we repeat the training data until it exceeds 2,000 for a stable convergence. Each object's point cloud has 10,000 points. In the experiments, we downsample 4,096, 1,024, 256, and 64 points in the four downsampling layers, respectively.

\begin{figure*}[t]
\centering
\includegraphics[width=0.9\linewidth]{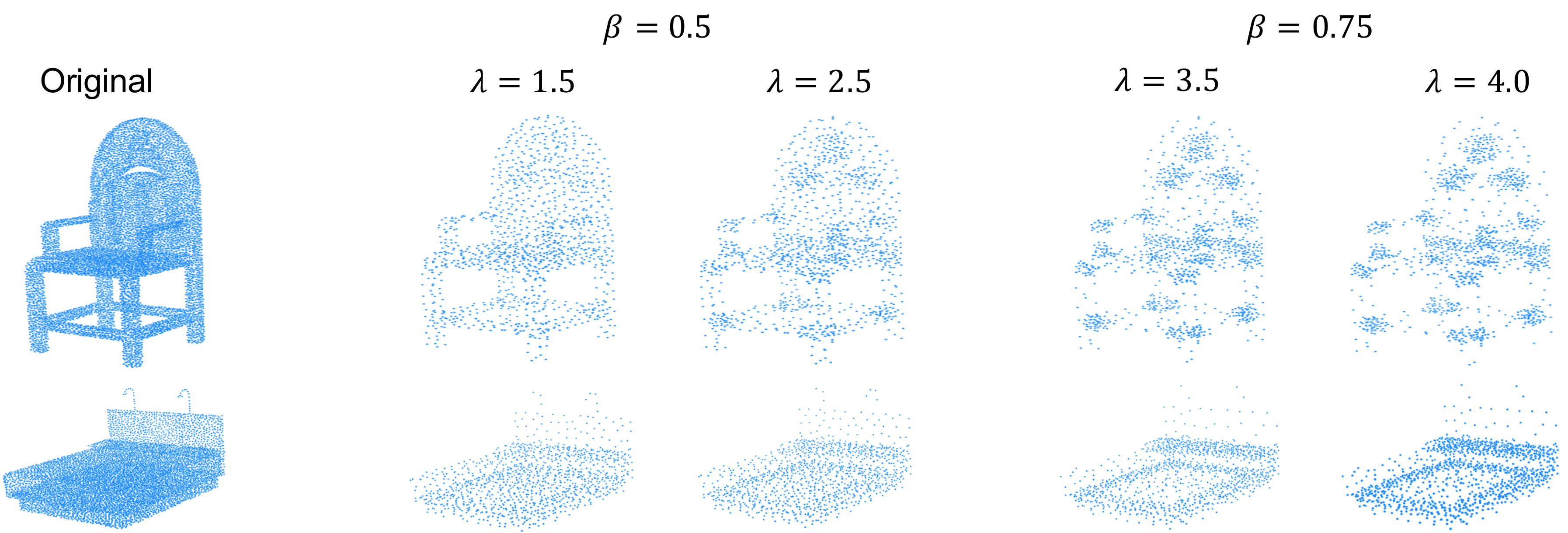}
\caption{Visualization results of \textit{Edge-FPS} with different hyperparameters ($\lambda$ and $\beta$). Points are sampled more densely in boundary areas than others. Larger $\lambda$ means to sample more points from boundary areas.}
\label{fig:edge_exps}
\end{figure*}

\vspace{0.8em}
\noindent\textbf{Evaluation metrics.}
We use the mean Intersection-over-Union (mIoU) scores as the evaluation metric. Following \cite{yu2019partnet}, we first remove the unlabeled ground truth points and calculate the IoU between the predicted points and ground truth points for each part category. Then, we obtain part-category mIoU by averaging the per-part-category IoU to evaluate the performance of every semantic part. We further calculate another evaluation metric, shape mIoU by averaging per-shape IoU to evaluate the performance on every object sample.

\vspace{0.8em}
\noindent\textbf{Design of supervising sampling.} We formally introduce \textit{Edge-FPS}. Firstly, we define a boundary point that  has at least one neighbor point (i.e., within a radius equals to $\epsilon$) labeled with a different semantic label \cite{loizou2020learning}. Let $\mathcal{P}_i$ be the input point set with $N_i$ points and $N_{i+1}$ be the number of points to sample. Supposing that the input point set \yq{can be split into boundary set $\mathcal{P}_i^b$ with $N_{i}^b$ points and non-boundary set $\mathcal{P}_i^n$ with $N_{i}^n$ points, where $N_i^b + N_i^n = N_i$,} the total number of boundary points to sample is given by:
\begin{equation} \label{eq:ege_fps}
    N_{i+1}^b = \text{max}(\lambda\frac{N_i^b}{N_i} N_{i+1}, N_i^b, \beta N_{i+1}),
\end{equation}
where $\lambda$ is the scale ratio $(\lambda \geq 1)$ and $\beta$ is the clip ratio $(0 \leq \beta \leq 1)$. The number of sampled non-boundary points is given by
\begin{equation}
    N_{i+1}^{n} = N_{i+1} - N_{i+1}^{b}.
\end{equation}
Thus, we propose \textit{Edge-FPS} to \yq{sample $N_{i+1}^{b}$ from boundary set $\mathcal{P}_i^b$ and $N_{i+1}^{n}$ points from non-boundary set $\mathcal{P}_i^n$ by FPS:
\begin{equation}
    \mathcal{P}_i' = 
    \text{FPS}(\mathcal{P}_i^b, N_{i+1}^{b}) \cup 
    \text{FPS}(\mathcal{P}_i^n, N_{i+1}^{n}).
\end{equation}}
As shown in TABLE \ref{tab:edge_fps}, \yq{we compare different $\lambda$ and $\beta$ in Equation~\ref{eq:ege_fps}. Compared with FPS, task-aware sampling (\textit{Edge-FPS}) brings a significant improvement of 4.9\% in shape mIoU and 5.2\% in part-category mIoU.} We choose the best setting of $\lambda = 3.5$ and $\beta = 0.75$ in the following experiments. Some visual examples are shown in Fig.~\ref{fig:edge_exps}. In addition to \textit{Edge-FPS}, an alternative sampling way is to sample points from each category point set respectively and more details can be referred to the comparison experiments in Section~\ref{sec:abla}.

\begin{table}
\begin{center}
\small
\renewcommand\tabcolsep{8pt}
\caption{Hyper-parameters ($\lambda$ and $\beta$) selection of \textit{Edge-FPS}. \yq{Experiments are conducted with PointNet++~\cite{qi2017pointnet++} in Chair-3.} A larger $\lambda$ means that there are more points sampled in boundary area. $\beta$ is a clip ratio to prevent sampling points from exceeding a certain percentage.} \vspace{-6pt}
\begin{tabular}{ll|c|c|c}
\toprule[1.5pt]
\multicolumn{2}{l|}{Method} & \multicolumn{1}{l|}{Shape mIoU} & \multicolumn{1}{l|}{Part mIoU} & \multicolumn{1}{l}{oA} \\ \hline \hline
\multicolumn{2}{l|}{Baseline} & 49.8 & 40.4 & 81.0 \\ \hline
$\beta=0.5$ & $\lambda=1.5$ & 52.0 & 44.6 & 84.2 \\
 & $\lambda=2$ & 53.4 & 43.4 & 84.7 \\
 & $\lambda=2.5$ & 54.0 & 44.6 & 85.2 \\
 & $\lambda=3$ & 54.2 & 45.3 & 85.6 \\ \hline
$\beta=0.75$ & $\lambda=3.5$ & \textbf{54.7} & \textbf{45.6} & \textbf{86.0} \\
 & $\lambda=4$ & 54.6 & 44.5 & 86.0 \\ \bottomrule[1.2pt]
\end{tabular}
\label{tab:edge_fps}
\end{center}
\end{table}

\begin{figure}[t]
\begin{center}
\includegraphics[width=0.95\linewidth]{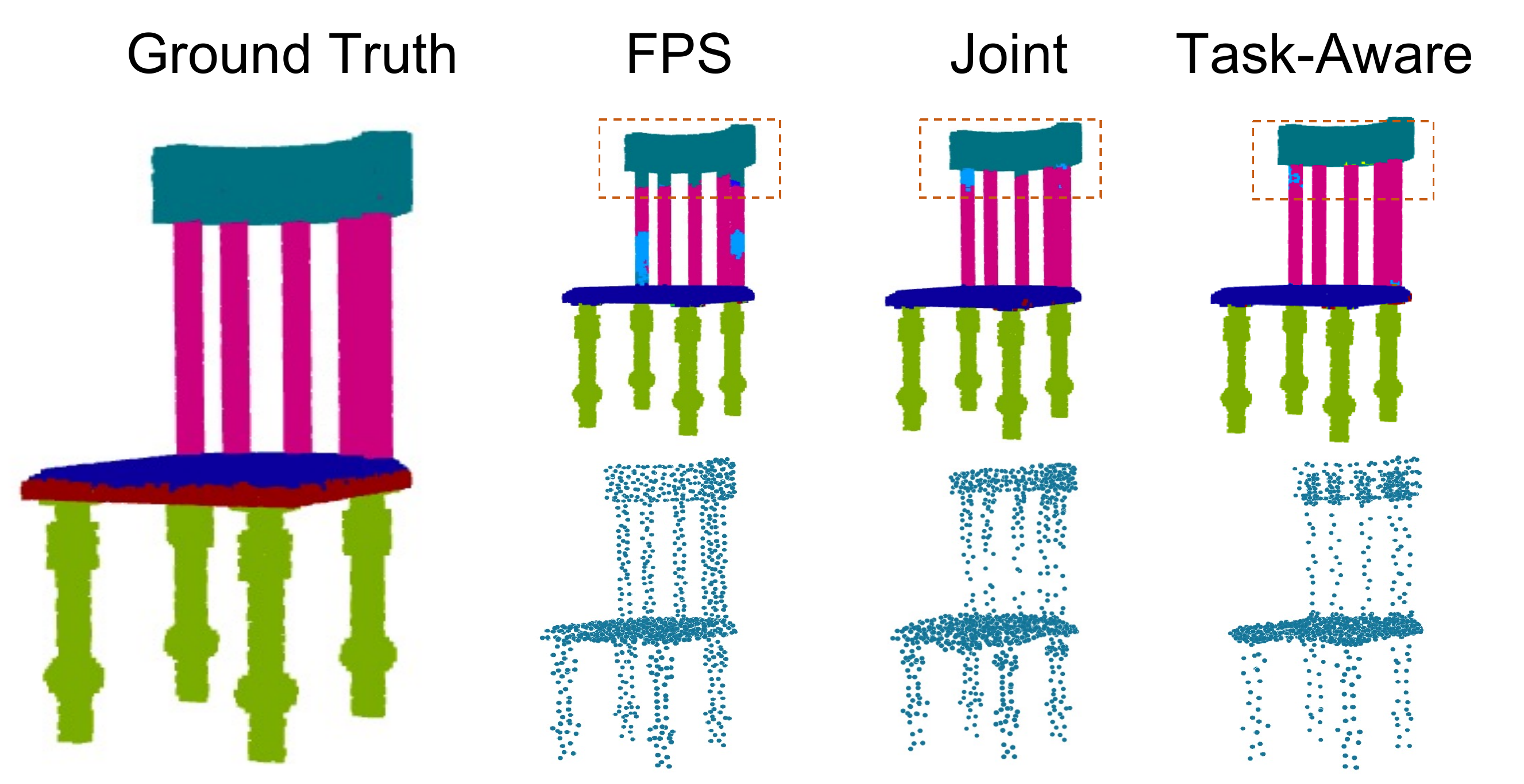}
\caption{Comparison of different sampling strategies (\textit{FPS}, \textit{Joint} and \textit{Task-Inspired} sampling) on semantic segmentation task. \textit{Joint} represents adopting learnable sampler that jointly supervised with task and displacement loss. The results show that predictions are more consistent near the boundary in task-aware sampling, and the jointly learned sampling can approximate it well.}
\label{fig:vis-seg}
\end{center}

\vspace{1.0em}
\begin{center}
\includegraphics[width=0.95\linewidth]{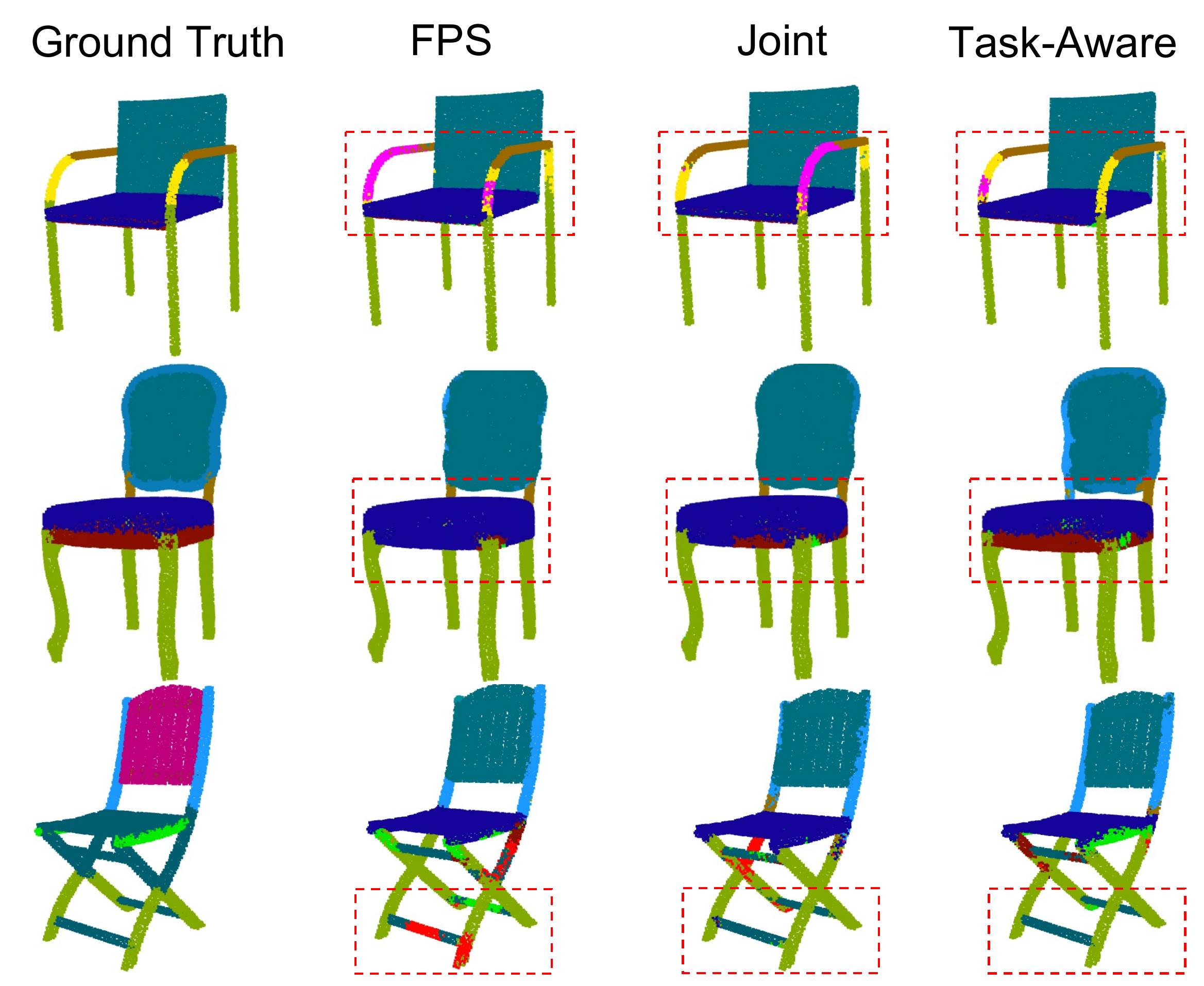}
\caption{Qualitative comparsion on PartNet~\cite{yu2019partnet} semantic segmentation.  Segmentation results near boundary areas are more accurate in \textit{Joint} and \textit{Task-Inspired} than in \textit{FPS}.}
\label{fig:visres_seg}
\end{center}
\end{figure}

\vspace{0.8em}
\noindent \textbf{Implementation.}
We use the cross-entropy loss as the task loss and train the whole network for 150 epochs for every category with a batch size of 12, optimized by momentum gradient descent optimizer with a momentum of 0.98 and an initial learning rate of 0.01. EMD$^\star$ loss function
in Equation \ref{eq:emd_hint} is used for jointly training. \yq{Point-wise offset learning strategy is adopted in the displacement network.} In addition, $\alpha$ in Equation \ref{eq:joint} is set to 50 initially and decayed by 0.95 for every 20 epochs. 
\yq{Note that we train a model separately for each category since the part semantics are different among categories (e.g., 57 parts in Chair and 23 parts in Clock).}

\vspace{0.8em}
\noindent \textbf{Results.}
As shown in TABLE~\ref{tab:part_seg_all}, in PointNet++ \cite{qi2017pointnet++} backbone, \yq{the task-aware sampling (\textit{Edge-FPS}; ``\textit{T}'' in TABLE~\ref{tab:part_seg_all}) on average brings a improvement of 2.3/2.2/2.5\% in shape mIoU and 3.9/1.8/1.6\% in part-category mIoU at three levels of segmentation, repectively.} The jointly learned sampling (``\textit{J}'' in TABLE~\ref{tab:part_seg_all}) on average outperforms the baseline model 0.7/1.1/1.6\% shape mIoU, and 1.4/1.1/1.4\% part-category mIoU.
\yq{In some categories such as Dish-1 and Vase-3, the improvements (i.e., 3.6/3.9 and 2.2/2.4, repsectively) are remarkable.} In Lamp-3 and Dishwasher-2, jointly learned sampling performs even better than using task-aware sampling.
%
From our point of view, the guidance from the task loss may be stronger than task-aware sampling points in such cases. However, in rare cases such as Hat-1 and Microwave-1, task-aware sampler performs even worse than FPS. It is possible that the information gained in boundary areas cannot compensate for the information loss caused by nonuniform sampling, and we will make an in-depth analysis in our future works. In addition, we find that the conclusions with other backbones including PointConv \cite{wu2019pointconv}, FPConv \cite{lin2020fpconv}, \yq{and PointCNN~\cite{li2018pointcnn}} are consistent, as shown in TABLE~\ref{tab:shape_partnet_full}. Visualization results are shown in Fig.~\ref{fig:vis-seg} and Fig.~\ref{fig:visres_seg}. 
Even equipped with the proposed sampling, the backbone models (i.e., PointNet++~\cite{qi2017pointnet++}, PointConv~\cite{wu2019pointconv}, FPConv~\cite{lin2020fpconv}, and PointCNN~\cite{li2018pointcnn}) still cannot exceed some previous state-of-the-art methods (e.g., MID-FC~\cite{Wang2020unsupervised}). We agree that a better backbone is important, and better training strategies like augmentation, pre-training, and advanced loss functions can further boost performance. However, few previous works focus on the study of sampling in point-wise analysis tasks. In this work, instead of lifting the performance to the state-of-the-art, we conducted a large number of experiments to demonstrate that a better sampling should also be a direction to improve the performance. In the future, there is potential to integrate our proposed sampling with more advanced backbones to achieve new state-of-the-art performance.

\begin{table*}[t]
\begin{center}
\footnotesize
\renewcommand\tabcolsep{3.1pt}
\caption{Semantic segmentation results (shape mIoU / part-category mIoU $\%$) on PartNet. $B$, $J$, and $T$ represent the baseline model, jointly learned sampling, and task-aware sampling respectively. The improvements of using jointly learned sampling are highlighted. The superscripts ($^{1, 2, 3}$) indicate different levels of segmentation: \yq{coarse-, middle- and fine-grained.} We do not show the results on category-levels that are not annotated in PartNet.} \label{tab:part_seg_all} \vspace{-6pt}
\scalebox{0.94}{
\begin{tabular}{c|ccccccccccccc}
\toprule[2.5pt]
  & \multicolumn{1}{c|}{\textbf{Avg.}}  & \textbf{Bag}  & \textbf{Bed}  & \textbf{Bottle}    & \textbf{Bowl} & \textbf{Chair}& \textbf{Clock}& \textbf{Dish.}    & \textbf{Disp.} & \textbf{Door} & \textbf{Earp.}& \textbf{Fauc.}& \textbf{Hat}  \\ \hline \hline 
\multicolumn{1}{c|}{$B^1$}  & \multicolumn{1}{c|}{70.9/61.2} & 66.3/57.0   & 49.8/55.7 & 82.5/\textbf{45.4} & 68.3/54.6 & 84.6/69.8 & \textbf{64.6}/43.0   & 64.2/69.0   & 90.8/91.1  & 70.0/64.5 & 71.9/\textbf{65.4} & 72.3/70.1 & \textbf{62.6}/\textbf{68.5} \\ \hline
\multicolumn{1}{c|}{$J^1$} & \multicolumn{1}{c|}{\begin{tabular}[c]{@{}l@{}}\textbf{71.6}/\textbf{61.9}\\ \red{+0.7}/\red{+0.7}\end{tabular}} & \begin{tabular}[c]{@{}l@{}}\textbf{67.5}/\textbf{57.5}\\ \red{+1.2}/\red{+0.5}\end{tabular} & \begin{tabular}[c]{@{}l@{}}\textbf{56.1}/\textbf{56.9}\\ \red{+5.3}/\red{+1.2}\end{tabular} & \begin{tabular}[c]{@{}l@{}}\textbf{83.4}/44.0\\ \red{+0.9}/\green{-1.4}\end{tabular} & \begin{tabular}[c]{@{}l@{}}\textbf{69.0}/\textbf{55.9}\\ \red{+0.7}/\red{+1.3}\end{tabular} & \begin{tabular}[c]{@{}l@{}}\textbf{85.1}/\textbf{70.0}\\ \red{+0.5}/\red{+0.2}\end{tabular} & \begin{tabular}[c]{@{}l@{}}64.5/\textbf{43.1}\\ \green{-0.1}/\red{+0.1}\end{tabular} & \begin{tabular}[c]{@{}l@{}}\textbf{67.8}/\textbf{72.9}\\ \red{+3.6}/\red{+3.9}\end{tabular} & \begin{tabular}[c]{@{}l@{}}\textbf{91.7}/\textbf{92.1}\\ \red{+0.9}/\red{+1.0}\end{tabular}  & \begin{tabular}[c]{@{}l@{}}\textbf{71.9}/\textbf{65.8}\\ \red{+1.9}/\red{+1.3}\end{tabular} & \begin{tabular}[c]{@{}l@{}}\textbf{73.2}/64.7\\ \red{+1.3}/\green{-0.7}\end{tabular} & \begin{tabular}[c]{@{}l@{}}\textbf{74.3}/\textbf{70.7}\\ \red{+2.0}/\red{+0.6}\end{tabular} & \begin{tabular}[c]{@{}l@{}}61.7/67.9\\ \green{-0.9}/\green{-0.6}\end{tabular} \\ \hline
\multicolumn{1}{c|}{$T^1$}  & \multicolumn{1}{c|}{73.2/64.4} & 69.8/60.8 & 60.1/58.3 & 83.7/47.2 & 71.2/58.6 & 89.1/81.0 & 65.4/45.9 & 69.8/75.2 & 95.4/94.9  & 76.9/71.1 & 75.2/65.3 & 75.4/70.8 & 59.5/67.4 \\ \midrule[2.0pt]
\multicolumn{1}{c|}{$B^2$}  & \multicolumn{1}{c|}{49.5/44.8} & - & 30.5/40.7 & - & - & 55.2/\textbf{45.6} & - & 53.4/55.8 & -  & \textbf{57.9}/53.1 & - & - & - \\ \hline
\multicolumn{1}{c|}{$J^2$} &\multicolumn{1}{c|}{\begin{tabular}[c]{@{}l@{}}\textbf{50.6}/\textbf{45.7}\\ \red{+1.1}/\red{+0.9}\end{tabular}} & - & \begin{tabular}[c]{@{}l@{}}\textbf{33.6}/\textbf{42.7}\\ \red{+3.1}/\red{+2.0}\end{tabular} & - & - & \begin{tabular}[c]{@{}l@{}}\textbf{55.4}/43.2\\ \red{+0.2}/\green{-2.4}\end{tabular} & - & \begin{tabular}[c]{@{}l@{}}\textbf{55.2}/\textbf{58.8}\\ \red{+1.8}/\red{+3.0}\end{tabular} & -  & \begin{tabular}[c]{@{}l@{}}55.7/\textbf{53.6}\\ \green{-2.2}/\red{+0.5}\end{tabular} & - & - & - \\ \hline
\multicolumn{1}{c|}{$T^2$}  & \multicolumn{1}{c|}{51.7/46.6} & - & 34.8/45.5 & - & - & 60.6/50.1 & - & 54.4/54.2 & -  & 59.1/54.6 & - & - & - \\ \midrule[2.0pt]
\multicolumn{1}{c|}{$B^3$}  &\multicolumn{1}{c|}{50.0/44.2}& - & 23.1/\textbf{39.4} & 62.1/42.3 & - & 49.8/40.4 & 40.9/31.6 & 48.0/47.6 & 81.3/81.7  & 47.5/41.0 & \textbf{51.0}/\textbf{43.6} & 59.3/61.7 & - \\ \hline
\multicolumn{1}{c|}{$J^3$} & \multicolumn{1}{c|}{\begin{tabular}[c]{@{}l@{}}\textbf{51.9}/\textbf{45.6}\\ \red{+1.9}/\red{+1.4}\end{tabular}} & - & \begin{tabular}[c]{@{}l@{}}\textbf{24.4}/38.6\\ \red{+1.3}/\green{-0.8}\end{tabular} & \begin{tabular}[c]{@{}l@{}}\textbf{63.5}/\textbf{43.4}\\ \red{+1.4}/\red{+1.1}\end{tabular} & - & \begin{tabular}[c]{@{}l@{}}\textbf{51.0}/\textbf{41.1}\\ \red{+1.2}/\red{+0.7}\end{tabular} & \begin{tabular}[c]{@{}l@{}}\textbf{41.0}/\textbf{32.1}\\ \red{+0.1}/\red{+0.5}\end{tabular} & \begin{tabular}[c]{@{}l@{}}\textbf{51.7}/\textbf{54.3}\\ \red{+3.7}/\red{+6.7}\end{tabular} & \begin{tabular}[c]{@{}l@{}}\textbf{81.9}/\textbf{82.2}\\ \red{+0.6}/\red{+0.5}\end{tabular}  & \begin{tabular}[c]{@{}l@{}}\textbf{48.3}/\textbf{41.5}\\ \red{+0.8}/\red{+0.5}\end{tabular} & \begin{tabular}[c]{@{}l@{}}50.6/41.9\\ \green{-0.4}/\green{-1.7}\end{tabular} & \begin{tabular}[c]{@{}l@{}}\textbf{62.6}/\textbf{62.9}\\ \red{+3.3}/\red{+1.2}\end{tabular} & - \\ \hline
\multicolumn{1}{c|}{$T^3$}  &\multicolumn{1}{c|}{52.5/45.8} & - & 25.5/42.0 & 67.7/44.5 & - & 54.2/44.0 & 42.5/33.4 & 47.4/51.1 & 85.5/85.7  & 46.7/40.6 & 50.7/39.7 & 63.2/60.9 & - \\ \midrule[2.5pt]
& \textbf{Keyb.}& \textbf{Knife}& \textbf{Lamp} & \textbf{Laptop}    & \textbf{Micro.}& \textbf{Mug}  & \textbf{Refri.}& \textbf{Scis.}& \textbf{Stora.}    & \textbf{Table}& \textbf{Trash.}    & \textbf{Vase} &   \\ \hline \hline
\multicolumn{1}{c|}{$B^1$}  & 67.8/67.5 & 63.0/53.4   & 37.5/28.6 & \textbf{95.8}/95.2 & \textbf{81.5}/\textbf{53.6} & 84.3/79.0   & 49.5/\textbf{43.6} & \textbf{82.1}/79.4 & 60.8/\textbf{62.6}  & 85.8/23.4 & 60.4/65.5 & 77.7/63.9 &   \\ \hline
\multicolumn{1}{c|}{$J^1$} & \begin{tabular}[c]{@{}l@{}}\textbf{68.4}/\textbf{68.4}\\ \red{+0.6}/\red{+0.9}\end{tabular} & \begin{tabular}[c]{@{}l@{}}\textbf{64.1}/\textbf{55.7}\\ \red{+1.1}/\red{+2.3}\end{tabular} & \begin{tabular}[c]{@{}l@{}}\textbf{39.8}/\textbf{30.5}\\ \red{+2.3}/\red{+1.9}\end{tabular} & \begin{tabular}[c]{@{}l@{}}95.7/\textbf{95.2}\\ \green{-0.1}/\red{+0.0}\end{tabular} & \begin{tabular}[c]{@{}l@{}}79.5/53.5\\ \green{-2.0}/\green{-0.1}\end{tabular} & \begin{tabular}[c]{@{}l@{}}\textbf{85.4}/\textbf{79.3}\\ \red{+1.1}/\red{+0.3}\end{tabular} & \begin{tabular}[c]{@{}l@{}}\textbf{51.8}/43.0\\ \red{+2.3}/\green{-0.6}\end{tabular} & \begin{tabular}[c]{@{}l@{}}81.7/\textbf{79.4}\\ \green{-0.4}/\red{+0.0}\end{tabular} & \begin{tabular}[c]{@{}l@{}}\textbf{61.8}/62.2\\ \red{+1.0}/\green{-0.4}\end{tabular} & \begin{tabular}[c]{@{}l@{}}\textbf{86.1}/\textbf{25.2}\\ \red{+0.3}/\red{+1.8}\end{tabular} & \begin{tabular}[c]{@{}l@{}}\textbf{60.4}/\textbf{65.5}\\ \red{+0.0}/\red{+0.0}\end{tabular} & \begin{tabular}[c]{@{}l@{}}\textbf{77.9}/\textbf{66.2}\\ \red{+0.2}/\red{+2.3}\end{tabular} &   \\ \hline
\multicolumn{1}{c|}{$T^1$}  & 70.1/70.1 & 67.6/58.6 & 37.2/29.3 & 96.3/95.5 & 79.4/53.8 & 83.0/79.7 & 49.4/43.6 & 85.3/83.6 & 67.4/73.6  & 90.2/25.7 & 59.0/66.3 & 81.1/68.2 &   \\ \midrule[2.0pt]
\multicolumn{1}{c|}{$B^2$}  & - & - & 33.7/\textbf{29.0} & - & 60.0/\textbf{52.2} & - & 54.8/44.4 & - & 52.6./56.3 & 47.5/25.7 & - & - &   \\ \hline
\multicolumn{1}{c|}{$J^2$} & - & - & \begin{tabular}[c]{@{}l@{}}\textbf{36.4}/28.9\\ \red{+2.7}/\green{-0.1}\end{tabular} & - & \begin{tabular}[c]{@{}l@{}}\textbf{62.0}/51.5\\ \red{+2.0}/\green{-0.7}\end{tabular} & - & \begin{tabular}[c]{@{}l@{}}\textbf{55.9}/\textbf{47.3}\\ \red{+0.9}/\red{+2.9}\end{tabular} & - & \begin{tabular}[c]{@{}l@{}}\textbf{53.5}/\textbf{56.4}\\ \red{+0.9}/\red{+0.1}\end{tabular}  & \begin{tabular}[c]{@{}l@{}}\textbf{47.9}/\textbf{28.7}\\ \red{+0.4}/\red{+3.0}\end{tabular} & - & - &   \\ \hline
\multicolumn{1}{c|}{$T^2$}  & - & - & 35.1/28.2 & - & 62.3/54.1 & - & 56.0/48.2 & - & 53.9/57.8  & 49.0/27.0 & - & - &   \\ \midrule[2.0pt]
\multicolumn{1}{c|}{$B^3$}  & - & 41.6/32.5 & 29.3/21.1 & - & 52.7/48.3 & - & 46.5/40.5 & - & 51.2/\textbf{47.6}  & 42.1/25.4 & 46.7/\textbf{53.5} & 76.5/53.7 &   \\ \hline
\multicolumn{1}{c|}{$J^3$} & - & \begin{tabular}[c]{@{}l@{}}\textbf{44.8}/\textbf{35.6}\\ \red{+3.2}/\red{+3.1}\end{tabular} & \begin{tabular}[c]{@{}l@{}}\textbf{32.6}/\textbf{21.3}\\ \red{+3.3}/\red{+0.2}\end{tabular} & - & \begin{tabular}[c]{@{}l@{}}\textbf{55.3}/\textbf{51.8}\\ \red{+2.6}/\red{+3.5}\end{tabular} & - & \begin{tabular}[c]{@{}l@{}}\textbf{50.5}/\textbf{47.3}\\ \red{+4.0}/\red{+6.8}\end{tabular} & - & \begin{tabular}[c]{@{}l@{}}\textbf{52.2}/47.3\\ \red{+1.1}/\green{-0.3}\end{tabular}  & \begin{tabular}[c]{@{}l@{}}\textbf{43.5}/\textbf{25.4}\\ \red{+1.4}/\red{+0.0}\end{tabular} & \begin{tabular}[c]{@{}l@{}}\textbf{49.8}/53.3\\ \red{+3.1}/\green{-0.2}\end{tabular} & \begin{tabular}[c]{@{}l@{}}\textbf{78.7}/\textbf{55.3}\\ \red{+2.2}/\red{+1.6}\end{tabular} &   \\ \hline
\multicolumn{1}{c|}{$T^3$}  & - & 48.3/41.7 & 30.2/21.3 & - & 54.1/48.0 & - & 51.4/43.8 & - & 52.2/48.7  & 44.1/25.4 & 50.5/53.4 & 78.6/55.0 &   \\ \bottomrule[2.5pt]
\end{tabular}
}
\end{center}
\end{table*}

\begin{table}[t]
\begin{center}
\small
\renewcommand\tabcolsep{1.2pt}
\caption{\yqq{Part-category} mIoU $\%$ on PartNet \cite{yu2019partnet} semantic segmentation. We compare FPS (Base.), jointly learned sampling (Joint), and task-aware sampling (T.A.) in different backbone networks, including PointNet++ (PN) \cite{qi2017pointnet++}, PointConv (PC) \cite{wu2019pointconv}, FPConv (FP) \cite{lin2020fpconv} and PointCNN (CN)~\cite{li2018pointcnn}. The numbers (1, 2, 3) that follow the category indicate different segmentation levels.} \vspace{-6pt}
\begin{tabular}{ll|c|cccccc}
\toprule[1.2pt]
&& \textbf{Avg.} & \textbf{Bed-1} & \textbf{Chair-3} & \textbf{Disp.-3} & \textbf{Door-1} & \textbf{Knife-1} & \textbf{Table-1} \\ \hline \hline
\multicolumn{1}{l|}{}   & Base.  & 53.2 & 55.7   & 40.4   & 81.7   & 64.5   & 53.4  & 23.4    \\
\multicolumn{1}{l|}{PN} & Joint  & \textbf{54.5} & \textbf{56.9}   & \textbf{41.1}   & \textbf{82.2}   & \textbf{65.8} & \textbf{55.7}  & \textbf{25.2}    \\ \cline{2-9}
\multicolumn{1}{l|}{}   & T.A. & 57.2 & 58.3   & 44.0   & 85.7   & 71.1 & 58.6  & 25.7    \\ \midrule[1.0pt]
\multicolumn{1}{l|}{}   & Base.  & 54.0 & 56.1 & 42.2 & 82.9 & \textbf{64.3} & 54.4 & 24.2 \\
\multicolumn{1}{l|}{PC} & Joint  & \textbf{54.7} & \textbf{56.5} & \textbf{43.3} & \textbf{83.3} & 64.1 & \textbf{55.9} & \textbf{25.2} \\ \cline{2-9}
\multicolumn{1}{l|}{}   & T.A. & 58.7 & 59.2 & 46.5 & 86.1 & 75.0 & 59.3 & 25.9 \\ \midrule[1.0pt]
\multicolumn{1}{l|}{}   & Base.  & 54.6 & 57.2 & 41.6 & 82.7 & 65.8 & 55.5 & 24.7 \\
\multicolumn{1}{l|}{FP} & Joint  & \textbf{56.6} & \textbf{57.7} & \textbf{44.5} & \textbf{84.5} & \textbf{70.0} & \textbf{56.9} & \textbf{25.7} \\ \cline{2-9}
\multicolumn{1}{l|}{}   & T.A. & 61.0 & 61.1 & 49.7 & 87.2 & 82.3 & 59.7 & 26.1 \\ \midrule[1.0pt]
\multicolumn{1}{l|}{}   & Base.  & 54.4 & 55.8 & 43.9 & 82.5 & 63.3 & 63.3 & 17.3 \\
\multicolumn{1}{l|}{CN} & Joint  & \textbf{55.4} & \textbf{56.3} & \textbf{44.8} & \textbf{82.9} & \textbf{64.4} & \textbf{64.4} & \textbf{19.4} \\ \cline{2-9}
\multicolumn{1}{l|}{}   & T.A. & 58.4 & 59.1 & 46.6 & 84.4 & 68.9 & 68.9 & 22.5 \\ \bottomrule[1.2pt]
\end{tabular}
\label{tab:shape_partnet_full}
\end{center}
\end{table}


%% file: subfiles/sec4_experiments/sec4_2_completion.tex
\subsection{Point Cloud Completion}
\noindent \textbf{Dataset and evaluation metrics.}
ShapeNetCore~\cite{shapenet} is used for our training. We adopt train/validation/test split from \cite{Yi16} with eight categories (i.e., airplane, bench, car, chair, lamp, rifle, table, and watercraft) and 20k+ objects in total. In data preparation, each object is normalized into a unit cube and 8 partial point clouds are obtained by back-projecting the rendered depth maps from different viewpoints. We uniformly sample 8,192 points from the surface of the partial point cloud and 8,192 points from the complete point cloud for task supervision. The number of downsampling points are the same as in segmentation task in each layer. We repeat the training data until it exceeds 15,000. In the evaluation, we adopt Chamfer distance~\cite{huang2020pfnet} to evaluate the completion results on surface points.

\begin{figure}[t]
\begin{center}
\includegraphics[width=1\linewidth]{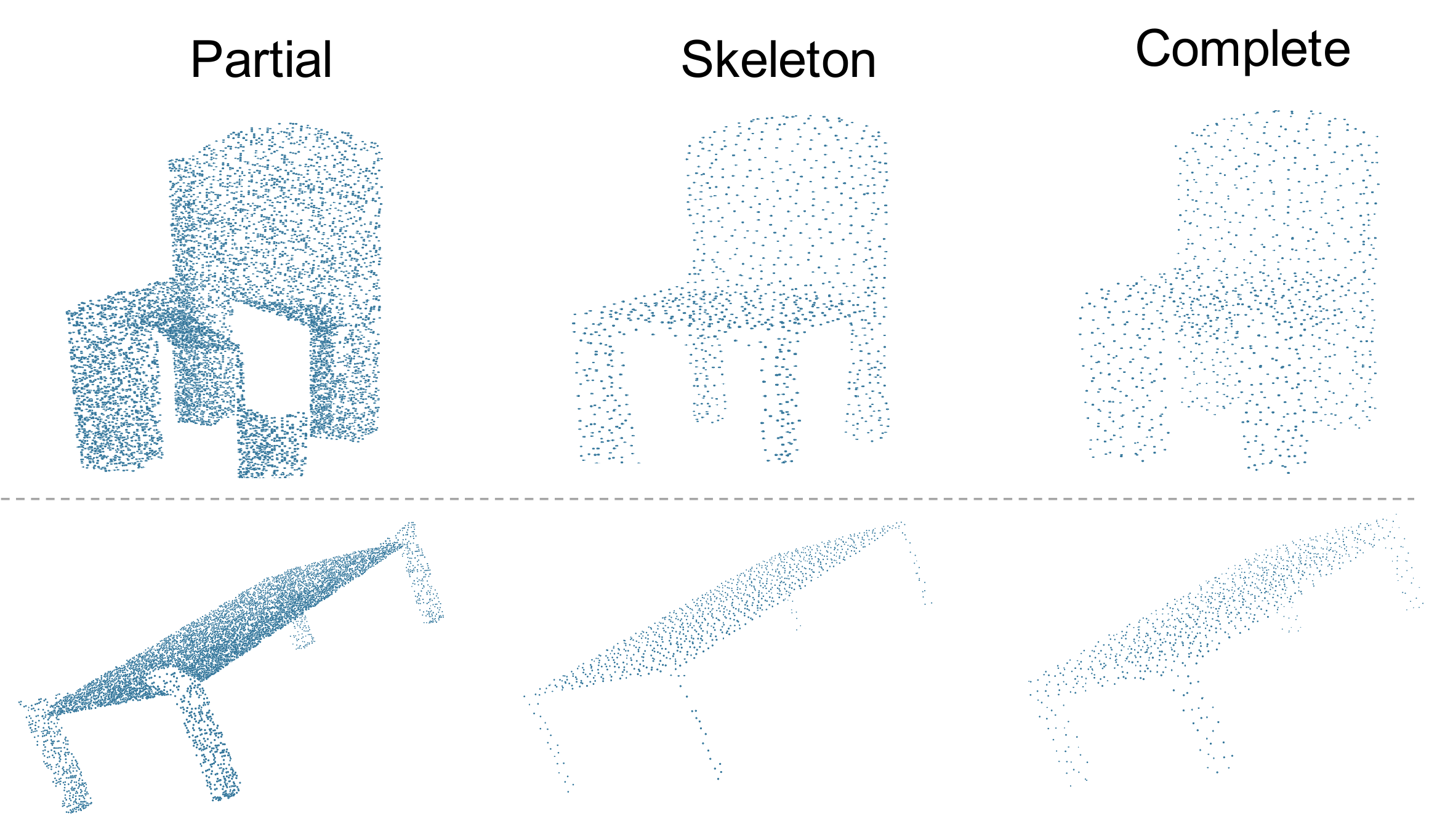}
\caption{Comparison of different task-aware samplings on point cloud completion. We compare sampling points from skeleton points (\textit{Skeleton}) \cite{Tang_2019_CVPR} and the complete point cloud (\textit{Complete}) for supervision.}
\label{fig:comp_exps}
\end{center}

\vspace{1.0em}
\begin{center}
\includegraphics[width=1\linewidth]{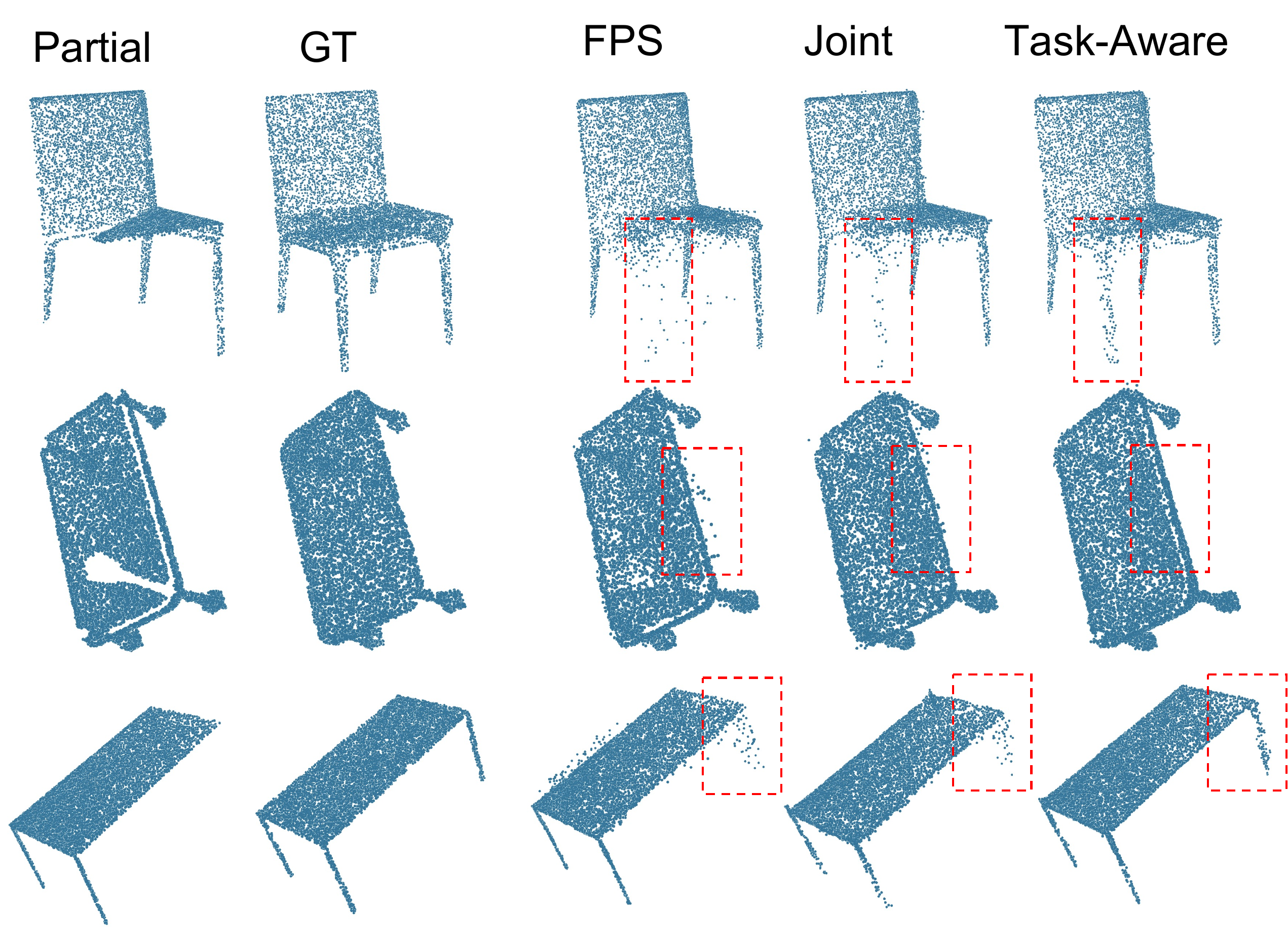}
\caption{Qualitative comparison on ShapeNet~\cite{shapenet} point cloud completion. Introducing ground truth information to the sampling phase makes the completion more accurate and reduces outlier points.}
\label{fig:resvis_pcc}
\end{center}
\end{figure}

\vspace{0.8em}
\noindent \textbf{Design of supervising sampling.} We downsample 1,024 points from the complete point cloud for sampling supervision. Another alternative sampling way is to sample from skeleton points \cite{Tang_2019_CVPR}. 
Comparison of two supervision samplings are shown in Section \ref{sec:abla} and we recommend directly sampling points from the complete point cloud for supervision since it is easy to obtain and has better performance. More visualization results are shown in Fig.~\ref{fig:comp_exps}.

\vspace{0.8em}
\noindent \textbf{Implementation.}
We train the networks for 60 epochs for every category with a batch size of 12 and the same optimizer as in semantic part segmentation task. Differently, we use CD loss given in Equation~\ref{eq:cd} \yq{as task loss} for shape supervision.
\yqq{We use EMD$^\star$ in Equation~\ref{eq:emd_hint} as sampling supervision loss function and use point-wise offset learning in the displacement network (sampling).}
In addition, $\alpha$ in Equation~\ref{eq:joint} is set to 0.5 and the joint loss is scaled by $10^2$ for better convergence. 

\begin{figure}[t]
\begin{center}
\includegraphics[width=1\linewidth]{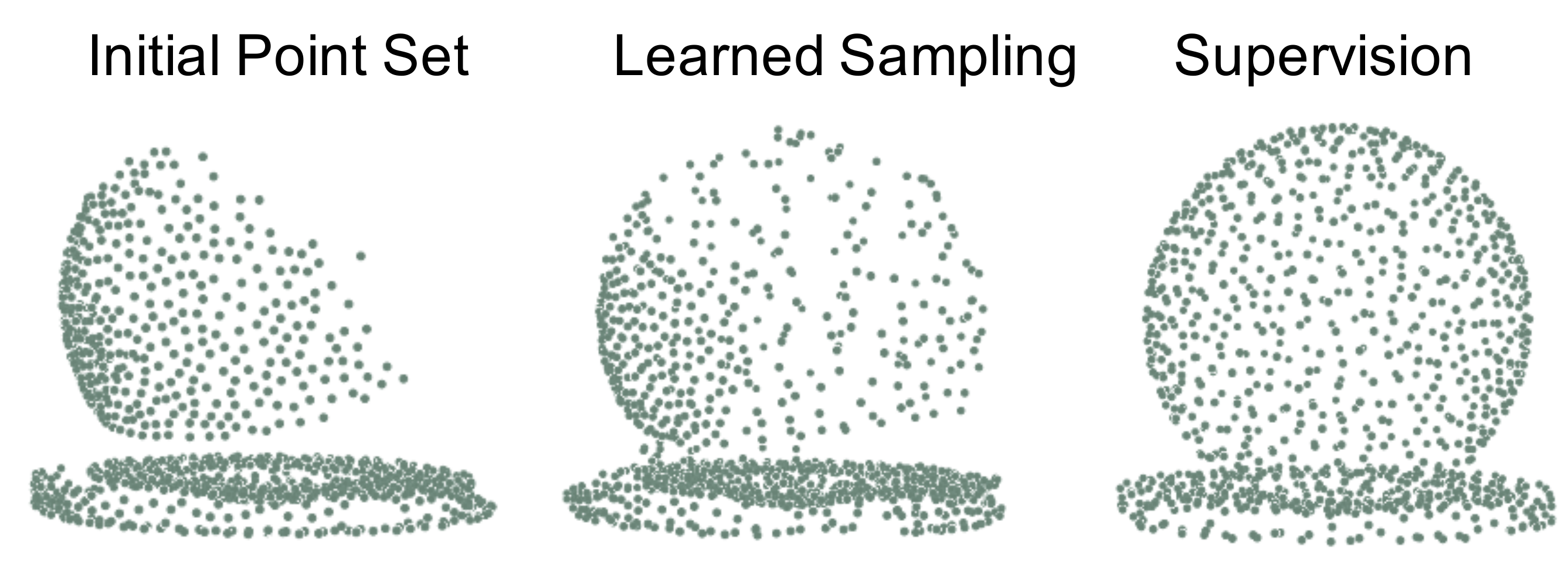}
\caption{Comparison of learned sampling with supervising sampling on point cloud completion task. Initial point set is sampled from the partial point cloud by FPS and supervised by the downsampled complete point cloud. we show that it is difficult to learn a point distribution that is too far away from the initial one.}
\label{fig:vis-pcc}
\end{center}
\end{figure}

\vspace{0.8em}
\noindent \textbf{Results.} As the experiments shown in TABLE~\ref{tab:pcc}, our method outperforms the baseline but still has a great gap with task-aware sampler. To analyze, we visualize the predicted sampling points in Fig.~\ref{fig:vis-pcc}. \yq{Because the complete point cloud and the partial point cloud have a big difference in the point distribution,}
%
it becomes more difficult to learn point displacement than in segmentation. Visualization results are shown in Fig.~\ref{fig:resvis_pcc}.

\begin{table}[t]
\begin{center}
\small
\renewcommand\tabcolsep{8.2pt}
\caption{Quantitative comparisons on completion task. We compare charmfer distance ($\times 10^3$) between predicted points and complete point cloud on test/val dataset splits. * indicates $\times 10^{4}$. \textit{Joint} indicates jointly training the sampler with task and displacement loss, and \textit{Task-Aware} presents the task-aware sampling.} \vspace{-6pt}
\begin{tabular}{l|cc|c}
\toprule[1.2pt]
(test/val) $\downarrow$ & Baseline    & Joint       & Task-Aware      \\ \hline \hline
\textbf{Airplane}*  & 0.586/0.337 & \textbf{0.573}/\textbf{0.336} & 0.455/0.332 \\ 
\textbf{Bench}      & 0.144/0.136 & \textbf{0.137}/\textbf{0.132} & 0.085/0.084 \\ 
\textbf{Car}        & \textbf{0.132}/\textbf{0.121} & 0.135/0.126 & 0.117/0.110 \\ 
\textbf{Chair}      & 0.192/0.177 & \textbf{0.184}/\textbf{0.171} & 0.123/0.116 \\ 
\textbf{Lamp}       & \textbf{0.178}/0.148 & 0.184/\textbf{0.147} & 0.103/0.085 \\ 
\textbf{Rifle}*     & 0.351/0.372 & \textbf{0.313}/\textbf{0.329} & 0.337/0.347 \\ 
\textbf{Table}      & 0.250/0.212 & \textbf{0.229}/\textbf{0.201} & 0.132/0.123 \\ 
\textbf{Watercraft} & \textbf{0.113}/\textbf{0.105} & 0.115/0.106 & 0.079/0.075 \\ \hline
\textbf{Avg.} & 0.138/0.121 & \textbf{0.134}/\textbf{0.118} & 0.090/0.083 \\ \bottomrule[1.2pt]
\end{tabular}
\label{tab:pcc}
\end{center}
\end{table}

%% file: subfiles/sec4_experiments/sec4_3_keypoint.tex
\subsection{Keypoint Detection}
\noindent \textbf{Dataset and evaluation metrics.}
We conduct experiments on KeypointNet~\cite{you2020keypointnet}, a large-scale and diverse 3D keypoint dataset that contains 103,450 keypoints and 8,234 3D objects from 16 categories. We compare different sampling strategies on four categories (i.e., bottle, chair, car, and table) with 2,048 input points. Following the data preparation of \cite{you2020keypointnet}, each object point cloud is normalized into a unit sphere. We downsample 1,024, 256, 64, and 16 points for multi-scale learning. The training data is repeated until it exceeds 2,000 in one epoch. A keypoint is considered as being detected when there is at least one predicted keypoint within a given distance. We evaluate the average precision (AP) of all object models within a distance threshold of 0.05.

\begin{figure}[t]
\begin{center}
\includegraphics[width=1\linewidth]{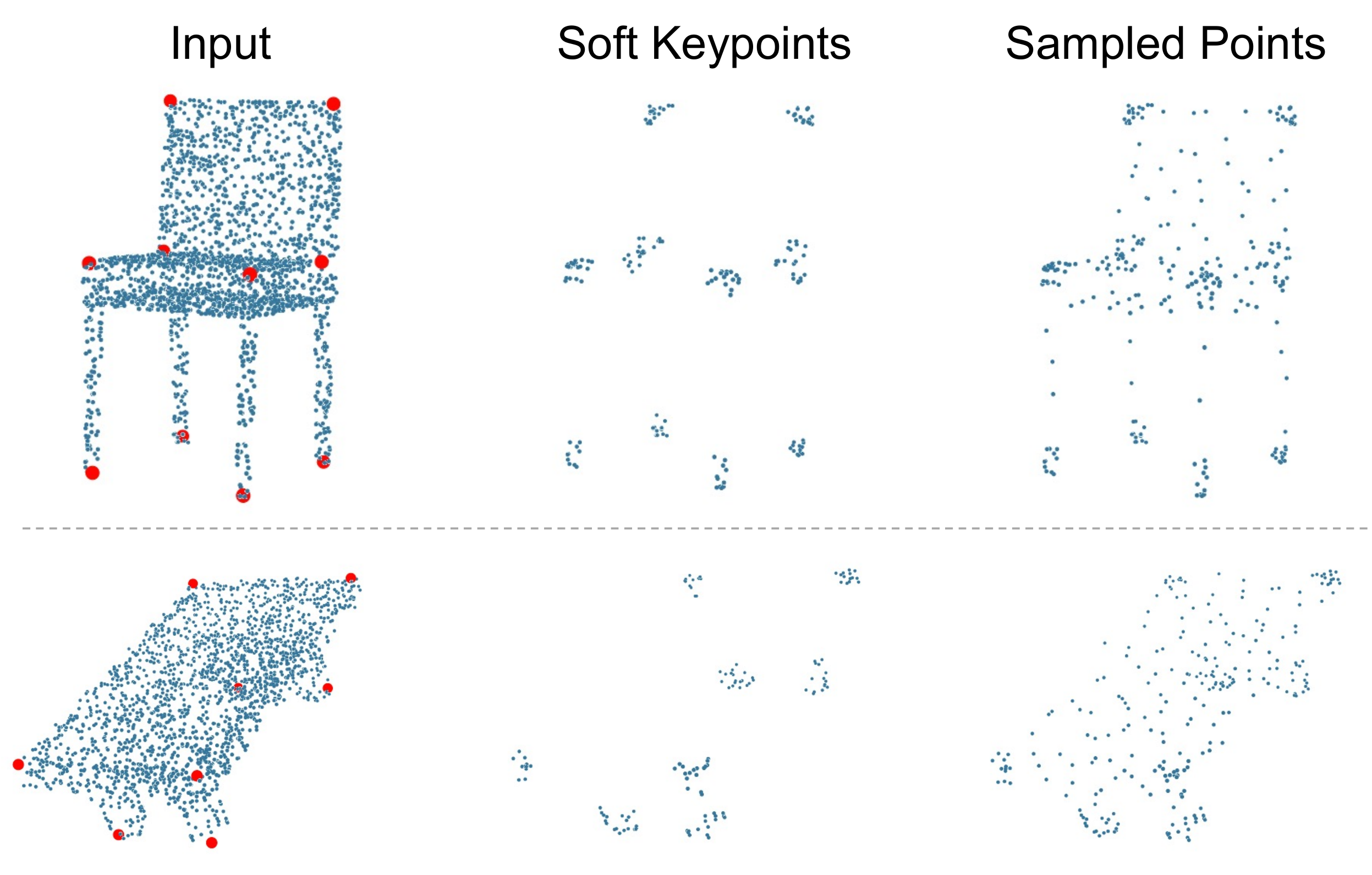}
\caption{Visual examples of task-aware sampling on keypoint detection. Soft keypoints are points near the labeled keypoints. Similar to \textit{Edge-FPS}, points are sampled more densely from soft keypoints and sparsely from other areas.}
\label{fig:keys_exps}
\end{center}

\vspace{1.0em}
\begin{center}
\includegraphics[width=1\linewidth]{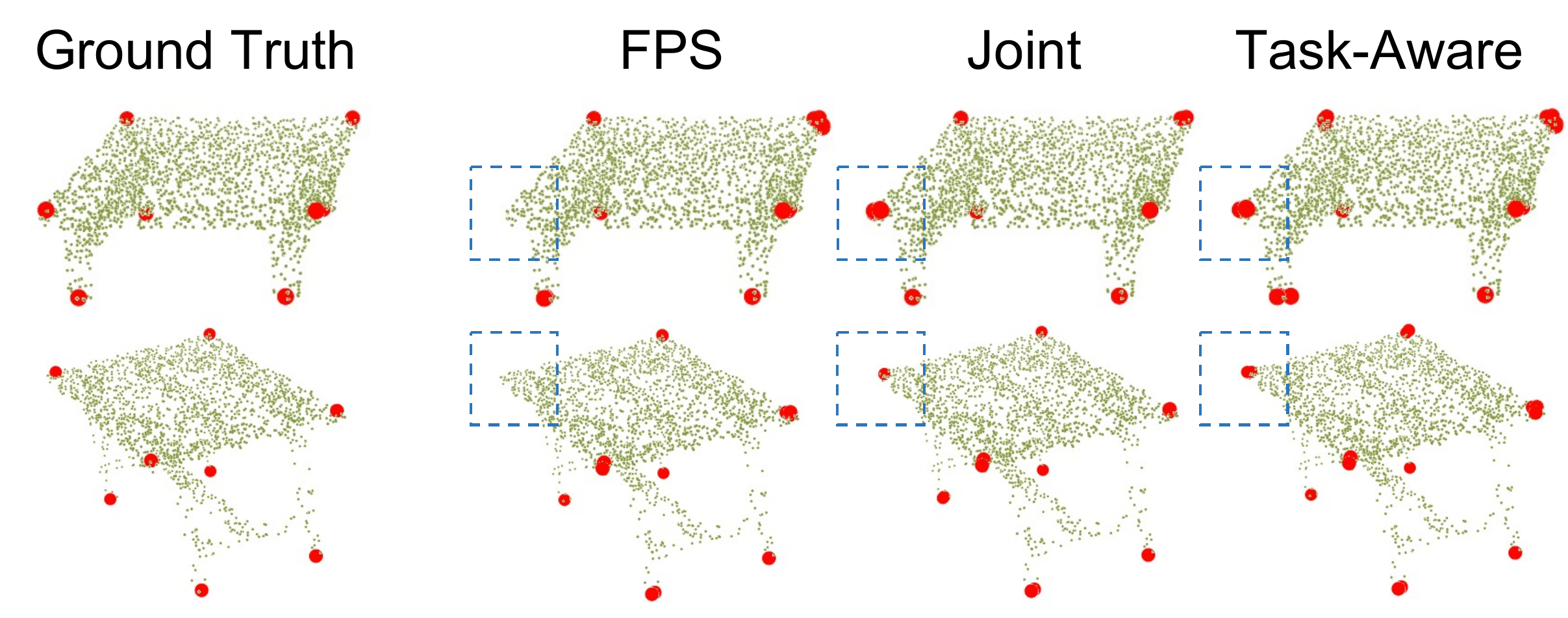}
\caption{Qualitative comparision on KeypointNet~\cite{you2020keypointnet} keypoint detection. Points in red are labeled or predicted (with a classification threshold of 0.5) keypoints. Sampling more points near keypoints (\textit{Joint} and \textit{Task-Inspired}) helps the task network identify keypoints more correctly.}
\label{fig:resvis_key}
\end{center}
\end{figure}

\vspace{0.8em}
\noindent \textbf{Design of supervising sampling.} Firstly, since the number of labeled keypoints is only 10 to 20 in an object, we regard these points that locate in 20-NN (Nearest Neighbors) of labeled keypoints as \textit{soft keypoints}. Similar to \textit{Edge-FPS}, we propose to sample more points from \textit{soft keypoints} and fewer from other areas as task-aware sampler. We set $\lambda=7.5$ and $\beta=0.8$ in Equation~\ref{eq:ege_fps} in following experiments. Visual examples are shown in Fig.~\ref{fig:keys_exps}.


\vspace{0.8em}
\noindent \textbf{Implementation and results.} Models are trained for 100 epochs with a batch size of 6 and optimized by Adam optimizer with a learning rate of 0.01. We use binary cross-entropy loss function for binary classification \yq{(i.e., keypoint and background)} and CD loss in Equation~\ref{eq:cd} for displacement supervision (sampling). Since there are only 10$\sim$20 keypoints in each object model, we scale the loss weight of positive labels by 10 \yq{to alleviate the problem of class imbalance.} $\alpha$ in Equation~\ref{eq:joint} is set to 0.5 in this task. Quantitative results are shown in TABLE~\ref{tab:keypoint}, and our supervised sampler outperforms the baseline in all categories. Visualization results are shown in Fig.~\ref{fig:resvis_key}.

\begin{table}[t]
\begin{center}
\small
\renewcommand\tabcolsep{8pt}
\caption{Quantitative comparisons on keypoint detection with different backbones including PointNet++ (PN)~\cite{qi2017pointnet++} and DGCNN (DG)~\cite{dgcnn}. We conduct experiments on four categories and compare average precision $\%$ with a distance threshold of 0.05 (AP$^{0.05}$). Task-Ins. represents the task-aware sampling.} \label{tab:keypoint} \vspace{-6pt}
\begin{tabular}{l|l|c|cccc}
\toprule[1.2pt]
& & \textbf{Avg.} & \textbf{Bottle} & \textbf{Chair} & \textbf{Car} & \textbf{Table}   \\ \hline\hline

\multicolumn{1}{l|}{}   & Base.  & 67.1 &  68.5 & 57.8 & 66.2 & 75.8  \\ 
\multicolumn{1}{l|}{PN} & Joint  & \textbf{67.8} &  \textbf{69.0} & \textbf{58.9} & \textbf{66.6} & \textbf{76.8} \\ \cline{2-7}
\multicolumn{1}{l|}{} & T.A.  & 70.4 & 74.8 & 59.3 & 68.9 & 78.5 \\ \midrule[1.0pt]

\multicolumn{1}{l|}{}   & Base.  & 68.1 & 64.7 &  64.1 & 63.4 & 80.3  \\ 
\multicolumn{1}{l|}{DG} & Joint  & \textbf{69.0} &  \textbf{65.2} & \textbf{65.7} & \textbf{64.1} & \textbf{80.9} \\ \cline{2-7}
\multicolumn{1}{l|}{} & T.A.  & 71.6 & 71.3 & 67.2 & 65.0 & 82.7 \\ \bottomrule[1.2pt]

\end{tabular}
\end{center}
\end{table}

%% file: subfiles/sec5_ablation.tex
\section{Ablation Study} \label{sec:abla}
In this section, we show several ablation studies on the sampler learning. Experiments are mainly conducted on the segmentation task of the Chair category at the segmentation level of 3 (PartNet \cite{yu2019partnet}), since Chair-3 has enough training data (3k+) and part categories.

\vspace{0.8em}
\noindent \textbf{Layer selection.} \label{sec:layer_selection}
We conduct ablative experiments on studying the effect of \yq{our proposed sampling strategy} in different downsampling layers. In TABLE~\ref{tab:layer_selection}, although adopting task-aware sampling in more layers is better, cumulative sampler learning brings more random noise in the initial training stage, making it difficult for point displacement networks in the later layers to converge. On the other hand, the sampled points are extremely dense in the first layer and sparse in the last two layers, which results in receptive fields being either too small or too large. Thus, we conduct all other comparison experiments only performing supervision in the second downsampling layer.

\begin{table}[t]
\begin{center}
\small
\renewcommand\tabcolsep{8.5pt}
\caption{Layer selection. We adopt task-aware sampling (Task-A.) or jointly learned sampling (Joint) in different downsampling layers. The evaluation results (shape, part-category mIoU and overall accuracy $\%$) show that adopting such samplings in the second layer only is the best.} \label{tab:layer_selection} \vspace{-6pt}
\begin{tabular}{l|cccc|c|c|c}
\toprule[1.5pt]
                              & 1  & 2 & 3 & 4    & Shape & Part & oA \\ \hline \hline
\multirow{4}{*}{Task-A.} & & $\surd$ &  &       & \textbf{54.2}       & 44.0 & 86.0        \\
                              & &   & $\surd$  &   & 51.1       & 42.3 &83.8      \\
                              & & $\surd$ & $\surd$    &       & 53.4       & \textbf{45.0} & 85.7        \\
                              & $\surd$ & $\surd$ & $\surd$  & $\surd$  & 53.3 & 44.2 & \textbf{86.5}      \\ \hline
\multirow{2}{*}{Joint}        & \multicolumn{1}{l}{} & $\surd$ & \multicolumn{1}{l}{} & \multicolumn{1}{l|}{}  & \textbf{51.0}  & \textbf{41.1} & \textbf{81.4}      \\
                              & \multicolumn{1}{l}{} & $\surd$ & $\surd$ & \multicolumn{1}{l|}{$\surd$} & 50.7       & 38.9 &81.3      \\ \bottomrule[1.2pt]
\end{tabular}
\end{center}

\vspace{0.5em}
\begin{center}
\small
\renewcommand\tabcolsep{10.7pt}
\caption{Sampler training strategies. We compare the impact of two components of joint loss function, i.e., task loss and displacement loss. The last row indicates further finetuning the sampler (trained without task loss) with joint loss (i.e., task and displacement loss).} \label{tab:joint_train} \vspace{2pt}
\begin{tabular}{l|c|c|c}
\toprule[1.5pt]
                                & Shape mIoU      & Part mIoU      & oA \\ \hline\hline
Baseline                        & 49.8       & 40.4      & 81.0         \\ \hline
Joint                     & \textbf{51.0} & \textbf{41.1} & \textbf{81.4} \\
Task loss only                   & 39.7       & 32.1      & 78.6         \\
w/o task loss                   & 49.5       & 40.3      & 81.0         \\
\ \ \ \ \ + finetune                      & 50.7       & 40.8      & 81.3     \\ \bottomrule[1.2pt]
\end{tabular}
\end{center}

\vspace{0.5em}
\begin{center}
\small
\renewcommand\tabcolsep{7.5pt}
\caption{Selection of task-aware samplers for supervision in semantic segmentation. The results show that good supervision is important for sampler learning, and \textit{Edge-FPS} is not the only choice.} \label{tab:sup_selection} \vspace{2pt}
\begin{tabular}{l|l|c|c|c}
\toprule[1.2pt]
Supervision & Loss & Shape mIoU & Part mIoU & oA \\ \hline \hline
- (baseline) & - & 49.8 & 40.4 & 81.0 \\ \hline
\multirow{2}{*}{FPS} & CD & 49.2 & 40.3 & 80.9 \\
 & EMD & 50.3 & 40.6 & 81.2 \\ \hline
Edge-FPS & EMD & \textbf{51.0} & \textbf{41.1} & \textbf{81.4} \\
Part-FPS & EMD & 50.9 & 40.4 & 81.3 \\ \bottomrule[1.2pt]
\end{tabular}
\end{center}

\vspace{0.5em}
\begin{center}
\small
\renewcommand\tabcolsep{6.7pt}
\caption{Selection of task-aware samplers in point cloud completion. We randomly downsample 1,024 points from complete point cloud and skeleton point cloud for supervision. Chamfer distance ($\times 10^3$) is used as the evaluation metric. $\star$ indicates $\times 10^4$.}
\label{tab:sup_comple} \vspace{2pt}
\begin{tabular}{l|ccccc}
 \toprule[1.2pt]
 & \textbf{Chair} & \textbf{Airplane}$^\star$ & \textbf{Car} & \textbf{Bench} & \textbf{Lamp} \\ \hline \hline
Baseline & 0.177 & 0.337 & 0.121 & 0.136 & 0.148 \\ \hline
Complete & \textbf{0.116} & 0.332 & \textbf{0.110} & \textbf{0.084} & \textbf{0.085} \\
Skeleton & 0.127 & \textbf{0.327} & 0.118 & 0.089 & 0.108 \\ \bottomrule[1.2pt]
\end{tabular}
\end{center}
\end{table}

\vspace{0.8em}
\noindent \textbf{Training strategy.}
We compare different sampler training strategies in TABLE~\ref{tab:joint_train}. Sampler trained only with task loss performs much worse than the baseline model. Because without the constraint of displacement loss, the sampled points will distribute very far away from the original and cannot capture good shape details. On the other hand, training the sampler only with the displacement loss (i.e., \yq{block the gradients from task loss to the sampler}) cannot bring improvement over the baseline model. However, if we further finetune the sampler with task loss, it can still achieve considerable improvement. In our view, task-aware sampling is not easy to learn, while task loss can bring semantic hints to promote better sampler learning.


\vspace{0.8em}
\noindent \textbf{Sampling supervision.}
We refer to sampling more points in boundary areas as \textit{Edge-FPS}. Besides, we design two other supervisions, sampling points with FPS in each part separately (named \textit{Part-FPS}) and uniformly sampling points in the original point cloud (named FPS). As shown in TABLE~\ref{tab:sup_selection}, both \textit{Edge-FPS} and \textit{Part-FPS} \yq{(with task-related information)} bring considerable improvements over the baseline model, while FPS \yq{(without task-related information)} cannot. The possible reason is that supervised with FPS will bring similar sampling distribution, and the extracted features have trivial differences from directly using FPS. To prove that, we replace the learned sampler (supervised by FPS sampled points) with FPS, and only suffer 0.1\% shape mIoU loss. Thus, we can conclude that effective supervision is important for sampler learning, and \textit{Edge-FPS} is not the only choice. We can also find the same conclusion in point cloud completion task from TABLE~\ref{tab:sup_comple}.

\vspace{0.8em}
\noindent \textbf{Initial point set.} \label{sec:init_selection}
We compare different initial point sets used in displacement learning. We first introduce \textit{pseudo Edge-FPS} by sampling more points in pseudo-boundary areas predicted by a boundary detection network and fewer in others. Then we adopt sampled points as the initial point set for displacement learning. \yq{The architecture of boundary detection network is the same as in Fig.~\ref{fig:net_achitecture} and supervised with binary cross-entropy loss.} As shown in TABLE~\ref{tab:init_pst}, using \textit{pseudo Edge-FPS} as the initial point set outperforms the baseline model but \yqq{slightly} worse than directly using FPS sampled points as the initial point set. \yqq{In addition, we observe that using randomly sampled points as the initial point set also brings comparable performance, revealing that the dispalcement network is not sensitive to the initial point set.}


\vspace{0.8em}
\noindent \textbf{Displacement learning.}
\label{sec:disp_learn}
As shown in Fig.~\ref{fig:sampler_pipeline}, there are two ways to obtain the sampled points by predicting coordinates directly or point-wise offsets to the initial point set. The experiment results are shown in TABLE~\ref{tab:disp_learn}. Directly regressing coordinates is more difficult than offset learning since the points may move too far away from the initial positions. Inspired by \cite{lang2020samplenet}, we adopt soft projection on the predicted points and project onto the original point cloud to prevent too much distribution change. In most cases, we recommend using offset learning or coordinate learning with soft projection. Therefore, we adopt offset learning in other segmentation comparison experiments. However, for the case that the target distribution is far away from the original (full scan and partial scan in completion task), regressing coordinates can also be considered.



\begin{table}[t]
\begin{center}
\small
\renewcommand\tabcolsep{9.2pt}
\caption{Selection of initial point sets: FPS, \yqq{randomly sampled,} and pseudo Edge-FPS. We first train a network for boundary points detection, then generate pseudo Edge-FPS by sampling more points in the predicted boundary areas and fewer in others areas.} \label{tab:init_pst} \vspace{-6pt}
\begin{tabular}{l|c|c|c}
\toprule[1.2pt]
Initial Points         & Shape mIoU & Part mIoU & oA \\ \hline \hline
- (baseline)           & 49.8       & 40.4      & 81.0         \\ \hline
FPS                    & \textbf{51.0} & 41.1 & \textbf{81.4} \\
\yqq{Random}        & 50.9 & \textbf{41.2} & 81.3 \\
Pseudo Edge-FPS        & 50.8 & 41.1 & 81.2       \\ \bottomrule[1.2pt]
\end{tabular}
\end{center}

\vspace{0.5em}
\begin{center}
\small
\renewcommand\tabcolsep{8.4pt}
\caption{Displacement learning strategies: point-wise offset learning (Offsets), coordinate learning (Coords.), and learning with soft projection~\cite{lang2020samplenet} (Coords. + soft proj.).} \vspace{2pt}
\begin{tabular}{l|c|c|c}
\toprule[1.2pt]
                   & Shape mIoU & Part mIoU & oA \\ \hline \hline
Baseline           & 49.8       & 40.4      & 81.0       \\ \hline
Offsets            & \textbf{51.0} & \textbf{41.1} & \textbf{81.4} \\
Coords.            & 49.8       & 39.8      & 81.1       \\
Coords. + soft proj.     & 50.8       & 41.0      & 81.3       \\ \bottomrule[1.2pt]
\end{tabular}
\label{tab:disp_learn}
\end{center}

\vspace{0.5em}
\begin{center}
\centering
\small
\renewcommand\tabcolsep{11.5pt}
\caption{Selection of displacement loss in sampler learning: CD and EMD. CD + EMD indicates that the loss function is composed of CD and EMD loss. The proportions are 0.3 and 0.7, respectively.}
\label{tab:loss_selection} \vspace{2pt}
\begin{tabular}{l|l|c|c|c}
\toprule[1.2pt]
 & $\loss_\text{disp}$ & Shape & Part & oA \\ \hline \hline
Baseline & - & 49.8 & 40.4 & 81.0 \\ \hline
\multirow{3}{*}{Joint} & CD & 50.2 & 40.2 & 81.1 \\
 & EMD & \textbf{51.0} & \textbf{41.1} & \textbf{81.4} \\
 & CD + EMD & 50.8 & 40.5 & 81.3 \\ \bottomrule[1.2pt]
\end{tabular}
\end{center}

\vspace{0.5em}
\begin{center}
\small
\renewcommand\tabcolsep{8.8pt}
\caption{Number of sampling points. We train the model with a learnable sampler that samples less than 1,024 points (i.e., 896 and 768) in the second downsampling layer.}
\label{tab:npts_sampler} \vspace{2pt}
\begin{tabular}{l|l|c|c|c}
\toprule[1.2pt]
 & Num. & Shape mIoU & Part mIoU & oA \\ \hline \hline
Baseline & 1,024 & 49.8 & 40.4 & 81.0 \\ \hline
\multirow{3}{*}{Joint} & 1,024 & \textbf{51.0} & \textbf{41.1} & \textbf{81.4} \\
 & 896 & 50.6 & 39.8 & 81.1 \\
 & 768 & 50.1 & 39.3 & 80.9 \\ \bottomrule[1.2pt]
\end{tabular}
\end{center}
\end{table}

\vspace{0.8em}
\noindent \textbf{Displacement loss.} \label{sec:disp_loss}
In Section \ref{sec:learn_sampling}, we introduce two displacement loss functions for shape supervision, i.e., Chamfer distance (CD) loss, and earth mover's distance (EMD) loss. As shown in TABLE~\ref{tab:loss_selection}, training with EMD loss outperforms CD loss in semantic segmentation. The reason may be that we still have points sampled in other areas, though it is sparse, limiting the movement range of the initial points. That is, in bidirectional nearest point matching, each point can only be matched to a nearby point. Therefore, the learned distribution will not be far away from the initial. In other tasks, we also conduct comparison experiments and choose the best loss function.

\vspace{0.8em}
\noindent \textbf{Number of sampling points.}
We explore the impact of the number of sampling points on the performance of the second downsampling layer. As shown in TABLE~\ref{tab:npts_sampler}, adopting the jointly learned sampler with fewer (896 and 768) sampling points could still achieve better results than sampling 1024 points by FPS.

%% file: subfiles/sec6_discussion.tex
\section{Discussion} \label{sec:discussion}

\subsection{Comparison with SampleNet \cite{lang2020samplenet}} \label{sec:dis_samplenet}
\noindent
Although both SampleNet \cite{lang2020samplenet} and this work are studying point sampling, they are designed for different application areas and have different learning and supervising ways. We elaborate on their differences in detail from three aspects.

\vspace{0.8em}
\noindent \textbf{(a) They are designed based on different perspectives of sampling.} We would first emphasize there are fundamental differences between the purposes of our algorithm and SampleNet \cite{lang2020samplenet}. The goal of sampling in \cite{lang2020samplenet} is to sample a sparse subset that can be processed more easily, to represent the original point cloud, and maintain the similar performance of downstream tasks. While ours is to sample an intermediate point set for neighbor point grouping and feature aggregation to boost the overall task performance.

\vspace{0.8em}
\noindent \textbf{(b) The design of the point displacement module is different.} \cite{lang2020samplenet} generates a set of sampled points using the global feature which is extracted from the original point cloud. In our method, we select an initial point set and predict offsets or coordinates with point-wise features for each initial point.

\vspace{0.8em}
\noindent \textbf{(c) Direct application of SampleNet \cite{lang2020samplenet} does not work well.} \yq{As shown in TABLE~\ref{tab:disc_samplenet}}, we have tried to directly adopt the sampler learning strategy and module of SampleNet \cite{lang2020samplenet} on point-wise analysis networks (i.e., supervised by the original point cloud \yq{and generate points from global features}) while it brings no improvement. This is because the learned sampled points are still distributed uniformly and almost have no differences from FPS. \yq{Although we use \textit{Edge-FPS} as the displacement supervision, the improvement is still limited since it is difficult to learn local point distributions from global features.} \yq{In addition, we should point out that 
1.) with soft projection (a module used in SampleNet), the sampler cannot learn a point distribution that has different shape/contours (e.g., the completion task) since it projects the predicted points to the input point cloud to gain a similar point distribution with input points.
2.) however, without soft projection, the predicted points of SampleNet are messy and perform very worse in point-wise tasks, because the points are generated from the global features and it is extremely difficult to learn local differences from the global information.}


\begin{table}[t]
\begin{center}
\small
\renewcommand\tabcolsep{8.1pt}
\caption{\yq{Comparison of our proposed method and SampleNet~\cite{lang2020samplenet}. Although SampleNet can bring significant improvement over FPS with fewer sampling points in classification and registration tasks, the improvement is limited in semantic part segmentation task even we use \textit{Edge-FPS} as the displacement supervision.}} \label{tab:disc_samplenet} \vspace{-6pt}
\begin{tabular}{l|l|c|c|c}
\toprule[1.2pt]
Sampler & Supervision & Shape & Part & oA \\ \hline \hline
FPS & - & 49.8 & 40.4 & 81.0 \\ 
Edge-FPS & - & 54.2 & 44.0 & 86.0 \\ \hline
\multirow{2}{*}{SampleNet~\cite{lang2020samplenet}} 
& Original pcd. & 49.8 & 40.3 & 80.9 \\
& Edge-FPS & 50.0 & 40.6 & 81.1 \\ \hline
Ours & Edge-FPS & \textbf{51.0} & \textbf{41.1} & \textbf{81.4} \\ 
\bottomrule[1.2pt]
\end{tabular}
\end{center}
\end{table}

\subsection{Limitations of Sampling Supervision}
\textbf{(a) The task-aware sampling is not easy to learn.} We find that the improvement of jointly learned sampling is limited in some categories of segmentation task. We have experimented with several methods to learn sampled points, but sometimes the learned distribution cannot capture all local details perfectly. There are two possible reasons, (1) we think a deeper network can help improve the sampler learning with sufficient GPU memory; (2) Chamfer Distance (CD) and Earth Mover's Distance (EMD) loss are not good measurements for local density distribution. These would be left as directions of our future work.

\vspace{0.8em}
\noindent
\textbf{(b) Our method indeed requires a pre-design of sampling supervision.} For the tasks of semantic part segmentation, keypoint detection, and point cloud completion as reported above, we actually did not put much effort into the design of the task-aware samplers. Most importantly, our experiments show that sampling is not trivial and better samplers bring considerable improvements in point-wise analysis tasks. Although the design of task-aware sampler is heuristic, we hope that the design and learning of samplers can become one of the future research directions of 3D point clouds.

\vspace{0.8em}
\noindent
\yq{\textbf{(c) Other known limitations:} the learned sampling cannot be used in other tasks/datasets; a good design of sampling supervision is required in our proposed method; the sampler cannot be used if a network does not involve the downsampling process. We will leave them in future work.}

\begin{table}[t]
\centering
\renewcommand\tabcolsep{14pt}
\caption{\yq{Experiments on classification (ShapeNet~\cite{shapenet}) and indoor scene segmentation (ScanNet~\cite{dai2017scannet}) tasks. For the classification task, Edge-FPS and Part-FPS are designed based on the part semantic from PartNet~\cite{yu2019partnet}; Key-FPS is designed based on the keypoints from KeypointNet~\cite{you2020keypointnet}.}} \label{tab:cls_seg} \vspace{-6pt}
\begin{tabular}{l|c|c}
\toprule[1.2pt]
Sampling      & ShapeNet (Acc.) & ScanNet (mIoU) \\ \hline \hline
FPS (baseline)& 90.7            & 60.0    \\ \hline
Edge-FPS      & 90.7            & 59.9    \\
Part-FPS      & 90.7            & 60.1    \\ 
Key-FPS       & 90.8            & -       \\ \bottomrule[1.2pt]
\end{tabular}
\end{table}

\subsection{Other Point-Based Tasks} \label{sec:dis_cls_seg}
Sampling is a commonly used strategy for saving computational costs in point-based networks. In this paper, we demonstrate that there exists a better sampler than FPS in some point-wise analysis tasks by involving task-related information in sampled points. We have also tried to adapt our methods in other point-based tasks, including classification (ShapeNet~\cite{uy2019revisiting}), and indoor large-scene segmentation (ScanNet~\cite{dai2017scannet}), as shown in TABLE~\ref{tab:cls_seg}.
%
%
For classification, the improvement is limited no matter what sampling is used. From our point of view, the sampling can affect local feature aggregation via changing neighbor points, while the classification result of an object almost depends on the global feature, which is not sensitive to local differences.
For scene segmentation, the improvement is also trivial, either using Edge-FPS or Part-FPS. A possible reason is that objects in the scene are usually distributed separately and have clear boundaries with each other, while the remaining difficulty is to identify the category of each object. 
\yqq{In addition, the performance is almost unchanged when replacing the original sampler (FPS, used in training) with Edge/Part-FPS in the testing stage, which reveals that the model is not sensitive to the above samplers in the task of large scene segmentation.}
We have not thought of other better samplers so far, and this would be left in our future work.


\subsection{Computational Cost}
We should point out that our proposed learnable sampler does not introduce too much computational cost in the mentioned tasks. To be more specific, the training parameters of the baseline model (PointNet++~\cite{qi2017pointnet++} backbone) are 0.87M, while the proposed sampler only brings 0.11M ($\sim$13\%) extra parameters; the training \yqq{and inference} time of the baseline model is about 6-8h and 67ms respectively in segmentation task with a single NVIDIA RTX 2080Ti GPU, while introducing the proposed sampler increases the training time by 0.5h ($\sim$8\%) \yqq{and the inference time by 10ms ($\sim$15\%).}

%% file: subfiles/sec7_conclusion.tex
\section{Conclusion}
In this work, we explore the impact of sampling in point cloud analysis based on deep neural networks. Our observation is that uniform sampling like FPS is not always the optimal choice. We conduct analytical experiments and demonstrate that by introducing task-related information into the sampling process, the network can be trained more effectively for point-wise feature learning. We further present a novel pipeline for supervised sampler learning and achieve \yqq{better} performance than FPS in various tasks with different backbones. Moreover, task-aware samplers are manually designed and might not be the best. These bring two potential future research problems in point-wise analysis: ``What is the best sampling?'' and ``How to further learn a better sampler?''.

\ifCLASSOPTIONcompsoc
  \section*{Acknowledgments}
\else
  \section*{Acknowledgment}
\fi

This work was supported in part by the Key Area R\&D Program of Guangdong Province with grant No. 2018B030338001, by the National Key R\&D Program of China with grant No. 2018YFB1800800, by Shenzhen Outstanding Talents Training Fund, and by Guangdong Research Project No. 2017ZT07X152.